%% file: main.tex
\documentclass[11pt]{article}

\usepackage{acl}          

\usepackage{times}
\usepackage{latexsym}
\usepackage[T1]{fontenc}
\usepackage[utf8]{inputenc}
\usepackage{microtype}
\usepackage{inconsolata}

\usepackage{booktabs}
\usepackage{tabularx}
\usepackage{graphicx}
\usepackage{amsmath}
\usepackage{amsfonts}
\usepackage{amssymb}
\usepackage{xcolor}
\usepackage{enumitem}
\usepackage{listings}
\usepackage[most]{tcolorbox}
\newtcolorbox{promptbox}[2][]{
    enhanced,
    colback=white,
    colframe=black,
    arc=3.5mm,
    boxrule=0.8pt,
    left=8pt, right=8pt, top=10pt, bottom=8pt,
    breakable,
    fontupper=\small,
    attach boxed title to top left={yshift=-\tcboxedtitleheight/2, xshift=3mm},
    boxed title style={
        colback=white,
        colframe=black,
        arc=1.5mm,
        boxrule=0.8pt,
        top=2pt, bottom=2pt, left=5pt, right=5pt
    },
    title={\textbf{#2}},
    coltitle=black,
    #1
}

\usepackage{todonotes}

\title{\textbf{HALTT: A HAte Lexicon Through Time}}

\author{
  \textbf{Dario Franzese}$^{1}$ \quad \textbf{Farah Benamara}$^{2}$ \quad \textbf{Véronique Moriceau}$^{2}$ \quad \textbf{Viviana Patti}$^{1}$ \\
  $^{1}$Dipartimento di Informatica, Universit\`{a} di Torino, Italy \\
  $^{2}$IRIT, Universit{\'e} de Toulouse, France \\
  \texttt{dario.franzese@edu.unito.it} \quad \texttt{farah.benamara@irit.fr} \\
  \texttt{veronique.moriceau@irit.fr} \quad
  \texttt{viviana.patti@unito.it}
}

\date{}

\begin{document}

\maketitle

\begin{abstract}
Automated hate speech detection degrades over time as everyday vocabulary is weaponized into covert slurs while retaining original benign meanings (\textit{polysemy retention}). Existing lexical semantic change detection (LSCD) frameworks struggle with this phenomenon due to a lack of standardized, sense-delineated diachronic resources. We propose a multi-stage framework to construct and validate a diachronic lexicon of semantically shifted hate speech terms. Filtering $>20{,}000$ candidates, our pipeline combines historical ($1770$--$1964$) and modern ($2001$--$2024$) corpus frequency grounding, masked definition-level scoring to mitigate LLM temporal bias, and a dual-persona reasoning prompt paradigm. We release \textbf{HALTT}, a grounded lexicon of 19 polysemous terms with literal and derogatory senses, analyze shift typologies, and establish evaluation protocols for downstream diachronic benchmarking.

\textbf{Warning}: \textit{This paper contains examples of potentially hateful content}.

\end{abstract}

\input{sections/01_introduction}
\input{sections/02_related_work}
\input{sections/03_methodology}
\input{sections/04_analysis_discussion}
\input{sections/06_conclusion}

\input{sections/07_limitations_ethics}


\bibliography{custom}

\appendix
\input{sections/08_appendix}

\end{document}

%% file: sections/01_introduction.tex
\section{Introduction}
\label{sec:intro}

Automated content moderation systems on digital platforms face a persistent operational vulnerability: the rapid evolution of online abusive vocabulary~\citep{waseem-hovy-2016-hateful,vidgen-etal-2019-challenges}. Malicious actors routinely evade keyword-based and statistical filtering mechanisms by adapting everyday vocabulary into veiled slurs, coded insults, and pejorative dog whistles~\citep{magu-luo-2018-determining,nobata-etal-2016-abusive}. When benign terms are systematically co-opted into hateful contexts, machine learning classifiers trained on static, historically situated snapshots suffer from severe temporal performance degradation~\citep{loureiro-etal-2022-tempowic}. For instance, \citet{florio-etal-2020-time} demonstrate that hate speech models experience sharp performance degradation (reporting substantial drops in macro $F_1$-score exceeding 10--15\%) when evaluated on data temporally distant from 
the training window, while recent studies by \citet{ali-etal-2025-evolving} and \citet{hegde-etal-2025-chronoberg} show that pre-constructed lexicons and foundation models consistently fail to adapt to evolving slurs and normative shifts across time.

While the field of Lexical Semantic Change Detection (LSCD) has developed sophisticated frameworks to identify diachronic sense alterations~\citep{periti-montanelli-2024-lexical,schlechtweg-etal-2026-lscd}, standard LSCD paradigms often assume population-wide sense replacement across distant historical eras. In contrast, lexical shifts in toxic domains typically manifest as \textit{polysemy retention}: a target word retains its conventional, benign denotation in general discourse while simultaneously acquiring a covert, highly localized derogatory sense within polarized online subcultures~\citep{bourgeade-etal-2023-multilingual,hoeken-etal-2023-towards-detecting,kolli-etal-2026-word}. Although contextualized models can theoretically disambiguate explicit meanings in clear contexts, adversarial weaponization frequently exploits deliberate rhetorical ambiguity, double entendres, and plausible deniability. Words such as \texttt{snowflake} or \texttt{ratchet} continue to designate meteorological and mechanical concepts in literal contexts, yet function as politically charged insults or misogynistic slurs in contemporary digital forums.

A primary obstacle to tracing such semantic transformations over extended temporal horizons is the scarcity of standardized, sense-annotated diachronic linguistic resources. Existing hate speech datasets are predominantly cross-sectional, collecting social media posts over narrow temporal windows of one to four years~\citep{davidson-etal-2017-automated,founta-etal-2018-large,borkan-etal-2019-nuanced,basile-etal-2019-semeval,mollas-etal-2022-ethos}. Consequently, they lack the historical depth required to establish empirical baselines of neutral usage prior to semantic weaponization. Conversely, historical diachronic benchmarks, such as SemEval-2020 Task 1 and LSCDiscovery~\citep{schlechtweg-etal-2020-semeval,d-zamora-reina-etal-2022-black}, focus on general language evolution (e.g., changes in technology or register) and rarely isolate adversarial or abusive sense acquisition.

To bridge this critical resource gap, this paper introduces a multi-stage LLM-driven methodological framework for the curation, induction, and validation of \textbf{HALLT}, the first \textbf{Diachronic Hate Speech Lexicon in English}.
Our contributions are threefold: \\

(1)~\textbf{A Multi-Stage Curation Pipeline:} a scalable, modular methodology combining corpus-grounded frequency filtering, masked definition analysis, and dual-persona LLM role-playing to isolate polysemous lexical semantic change without human-in-the-loop candidate generation bias, easily adaptable to other languages;

(2)~\textbf{A Standardized Sense-Annotated Lexicon:} a curated 19-term benchmark (whose size is in line with related works in LSCD such as SemEval-2027 Task~3~\citep{tahmasebi-etal-2027-semeval}) equipped with an explicit set of definitions comprising 117 senses in total (distributed across 73 neutral and 44 derogatory or hateful senses), part-of-speech metadata, and longitudinal empirical distributions spanning over 50 million historical newspaper articles ($1770$--$1964$) and over 46 million contemporary posts aggregated across seven modern online datasets ($2001$--$2024$); 

(3)~\textbf{Quantitative and Qualitative Analysis:} an empirical examination of corpus frequency distributions confirming polysemy retention, a qualitative typology of pejorative shift mechanisms, and a multi-model consensus evaluation quantifying pipeline stability across stochastic LLM iterations.

%% file: sections/02_related_work.tex
\section{Related Work}
\label{sec:related_work}
\subsection{Static Hate Speech Lexicons}
Automated hate speech detection relies heavily on annotated benchmarks and domain-specific lexicons~\citep{schmidt-wiegand-2017-survey,poletto-etal-2021-resources}. Widely used lexical resources include \textit{HurtLex}~\citep{bassignana-etal-2018-hurtlex}, a multilingual lexicon of words to hurt categorized into semantic macro-categories; \textit{Hatebase}~\citep{hatebase-2022-hatebase}, a crowd-sourced repository of discriminatory terms; the \textit{Harassment Lexicon}~\citep{rezvan-etal-2018-quality}; and domain-targeted collections such as \textit{Taboo WordNet}~\citep{bond-choo-2021-taboo} and \textit{Abusive Words}~\citep{wiegand-etal-2018-inducing}. 

While these lexicons are invaluable for feature engineering and rule-based filtering, they are fundamentally static. Most existing repositories collapse lexical entries into uncontextualized lists or coarse binary tags without modeling how senses evolve over time or interact with non-offensive denotations. When deployed across longitudinal corpora, static wordlists suffer from high false positive rates due to benign polysemic uses. For example, \citet{davidson-etal-2017-automated} demonstrated that keyword matching on social media conflates general offensive colloquialisms with actual hate speech, yielding false-positive rates as high as 76\% when differentiating between hate and non-hateful offensive language. Similarly, \citet{wiegand-etal-2019-detection} found that lexical bias causes models to over-rely on surface identity tokens, generating excessive false alarms on benign occurrences. Recent studies by \citet{ali-etal-2025-evolving} further show that static keyword filters fail to adapt to evolving slurs and tactical spelling variations, while \citet{qureshi-etal-2026-training} and \citet{frenda-etal-2023-killing} stress the importance of grounding model reasoning in contextual and sociocultural rationales rather than rigid dictionary entries.

\subsection{Lexical Semantic Change Detection}
Computational modeling of lexical semantic change has progressed from diachronic static word embeddings~\citep{hamilton-etal-2016-diachronic,dubossarsky-etal-2019-time} to contextualized representations using transformer architectures~\citep{giulianelli-etal-2020-analysing,montariol-etal-2021-scalable,laicher-etal-2021-explaining,periti-montanelli-2024-lexical}. Standard evaluation frameworks, such as SemEval-2020 Task 1~\citep{schlechtweg-etal-2020-semeval}, LSCDiscovery~\citep{d-zamora-reina-etal-2022-black}, and the unified LSCD Benchmark~\citep{schlechtweg-etal-2026-lscd}, focus on measuring global degree of semantic change or binary sense divergence across historical eras. 

Recent advances have leveraged Large Language Models for automated definition generation and sense induction~\citep{fedorova-etal-2024-definition,fedorova-etal-2025-explaining,zhang-etal-2025-llm}, open-vocabulary sense retrieval for historical texts~\citep{ciletti-2026-inspicio}, Frame Semantics for interpretable change detection~\citep{tat-etal-2026-reframe}, multidimensional linguistic feature toolkits~\citep{phan-tat-etal-2026-synflow}, and word-centered semantic graph representations for tracking polysemy dynamics~\citep{kolli-etal-2026-word}. Additionally, large-scale multilingual diachronic corpora such as DHPLT~\citep{fedorova-etal-2026-dhplt} and temporally segmented pretraining architectures~\citep{fittschen-etal-2025-pretraining} have expanded evaluation horizons across multiple continuous epochs~\citep{kiyama-etal-2025-analyzing,jayatilleke-de-silva-2026-dynamics,liu-etal-2026-beyond}.

However, conventional LSCD benchmarks are designed primarily around general historical language (e.g., digitized newspapers from the 19th and 20th centuries) and lack explicit annotations for abusive or derogatory connotations. Consequently, standard LSCD models are rarely evaluated on adversarial semantic drifts where words acquire or lose weaponized meaning.

\subsection{Diachronic Hate and Concept Drift}
The intersection of diachronic linguistics and abusive language detection has garnered increasing attention due to the phenomenon of temporal degradation.
Studies investigating temporal robustness~\citep{florio-etal-2020-time,loureiro-etal-2022-tempowic}
consistently show that hate speech classifiers experience sharp performance drops when evaluated on data temporally distant from their training window. Furthermore, longitudinal analyses of social discourse reveal that ideological polarization drives continuous semantic divergence in conceptual framing~\citep{yitzchak-etal-2026-words}, while foundation models trained across long-term historical corpora struggle to capture evolving norms and time-sensitive discriminatory language~\citep{hegde-etal-2025-chronoberg}.

Pioneering efforts by~\citet{hoeken-etal-2023-towards-detecting} explored semantic change in 16th-century religious polemics, combining LSCD and toxicity annotations. 

Overall, existing resources in online toxicity either lack longitudinal breadth (covering only modern social media) or focus on historical epochs disconnected from contemporary digital hate subcultures. Our work addresses this gap by providing a principled methodology to construct a sense-annotated diachronic lexicon spanning over a century of usage.

%% file: sections/03_methodology.tex
\section{Lexicon Induction Methodology}
\label{sec:methodology}

\begin{figure*}[t]
\centering
\includegraphics[width=0.88\textwidth]{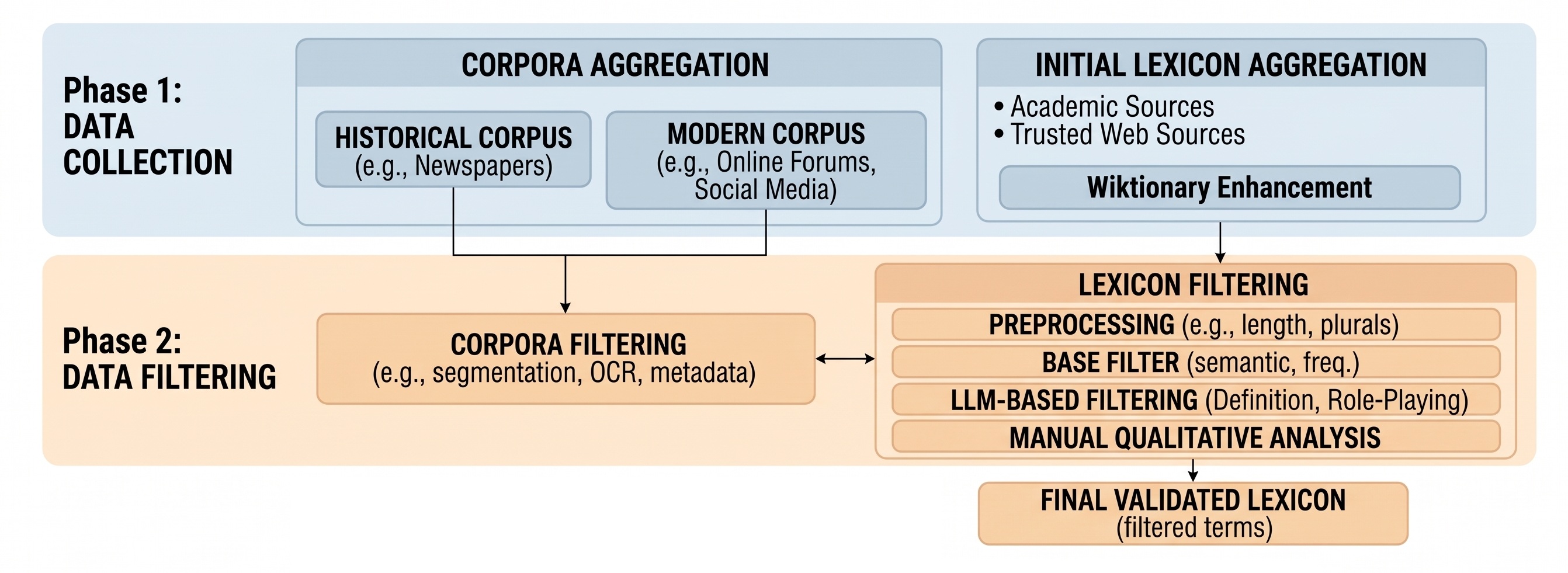}
\caption{The multi-stage lexicon construction and filtering pipeline. Starting from $>20,000$ candidate entries, the framework applies structural preprocessing, corpus grounding, masked LLM definition scoring, dual-persona role-playing, and qualitative sanity checks to isolate the final 19-term polysemous benchmark.}
\label{fig:pipeline}
\end{figure*}

The central objective of this research is the construction and validation of a diachronic hate speech lexicon characterized by semantic shift and polysemy retention. In this framework, large-scale textual corpora are utilized strictly as operational tools and empirical instruments to ground candidate words, enforce occurrence bounds, and provide authentic usage contexts, rather than serving as the primary artifact of this paper. The distillation pipeline progresses through five systematic stages, moving from broad candidate harvesting to granular qualitative validation. Figure~\ref{fig:pipeline} summarizes the end-to-end workflow.

\subsection{Lexicon Aggregation and Enhancement}
\label{subsec:candidate_aggregation}

\paragraph{Initial Candidate Aggregation.}
To capture a broad initial pool of abusive language and potentially shifted words, we aggregated nine established academic and trusted web repositories: HurtLex~\citep{bassignana-etal-2018-hurtlex}, Harassment Lexicon~\citep{rezvan-etal-2018-quality}, Abusive Words~\citep{wiegand-etal-2018-inducing}, Taboo WordNet~\citep{bond-choo-2021-taboo}, Google Profanity~\citep{gabriel-2018-google}, Hatebase~\citep{hatebase-2022-hatebase}, Weaponized Words~\citep{weaponized-word-2023-weaponized}, Wiktionary Derogatory Lists~\citep{wikimedia-foundation-2023-wiktionary}, and Wordnik Offensive Words~\citep{wordnik-2024-wordnik}. These sources were selected as they represent standard reference benchmark lexicons in abusive language detection. Their union yielded an initial pool of \textbf{21,263 raw candidate entries}. However, this raw aggregation carries substantial noise: many entries represent profanity, ad-hoc vulgarities, or highly context-dependent insults that lack longitudinal depth. A detailed overview of the source lexicons, their composition, raw sizes, and linguistic categories is reported in Table~\ref{tab:lexicons_summary} (Appendix~\ref{sec:appendix_lexicons_corpora}).

\paragraph{Lexicon Enhancement via Wiktionary.}
Standard wordlists treat entries as monolithic strings without differentiating word senses. To explicitly model polysemy, we scraped full Wiktionary entry hierarchies for all candidate terms using a custom scraper. Each entry was structured into discrete definitions containing: (1) textual definition glosses, (2) categorical domain and offensive register tags (e.g., \textit{derogatory} and \textit{slur}), (3) grammatical part-of-speech, and (4) illustrative quotations with timestamps. Although we had this temporal information, it was not directly used both because the timestamps were very sparse and to avoid bias towards the resource when relying on quotations for filters. This definition-level structuring enables fine-grained semantic reasoning, allowing subsequent filtering to operate over individual senses rather than ambiguous surface forms.

\subsection{Empirical Grounding Corpora}
\label{subsec:corpora_grounding}

To ground candidate words in authentic linguistic usage across time, we construct two massive textual archives as empirical testbeds. A comprehensive summary of all constituent corpora, time spans, volumes, and sampling properties is provided in Table~\ref{tab:corpora_summary} (Appendix~\ref{sec:appendix_lexicons_corpora}).

\paragraph{Historical Corpus ($\mathcal{C}_{\text{old}}$).}
We utilize \textit{American Stories}~\citep{dell-etal-2023-american}, an optical character recognition (OCR) archive comprising public-domain digitized newspapers from the Library of Congress spanning 1770 to 1964 (2023 snapshot). Due to the massive scale and historical OCR noise, which severely degrade specialized neural toolkits, articles were segmented into sentences using punctuation-based period-delimited boundaries, yielding approximately 1.08 billion sentences. Contextual usage instances for target terms were subsequently extracted using a 3-sentence sliding window (the target sentence flanked by its immediate predecessor and successor) to preserve essential surrounding discourse. This resource provides longitudinal depth back to the late 18th century with publication year metadata. As discussed in Section~\ref{sec:limitations}, the journalistic nature of this corpus imposes an intrinsic methodological constraint: editorial standards and linguistic restraint restrict overt offensive language primarily to rare direct quotations, which naturally predisposes observable diachronic trajectories toward shifts from non-offensive to offensive usages.

\paragraph{Unified Modern Corpus ($\mathcal{C}_{\text{modern}}$).}
Rather than relying on a single source, our modern corpus fuses two complementary components to capture contemporary toxic vernacular alongside diverse demographic targets. First, we fused six standardized online datasets~\citep{borkan-etal-2019-nuanced,toraman-etal-2022-large,chakraborty-2023-normal,jikeli-etal-2023-hate,kennedy-etal-2020-gab,kaggle-user-2019-hatred}, contributing 2.95 million instances. Second, we incorporated \textit{ExtremeBB}~\citep{vu-etal-2023-extremebb}, an unannotated repository of $\sim$50 million posts (2001--2024) from extremist online forums that provides substantial volume and high-density contemporary slang, slurs, and dog whistles. These datasets were chosen under the specific requisite that they constitute diachronic hate-labeled datasets, providing verified longitudinal timestamps and explicit abusive language annotations across diverse target groups. Fusing these sources yields an aggregated modern repository of over 52 million posts spanning 2001 to 2024 (detailed individually in Table~\ref{tab:corpora_summary}, Appendix~\ref{sec:appendix_lexicons_corpora}), mitigating scale imbalances against the historical archive. Sentence boundary disambiguation on contemporary text was conducted using PySBD~\citep{sadvilkar-neumann-2020-pysbd}.

\begin{figure*}[t]
\centering
\begin{minipage}[b]{0.48\textwidth}
    \centering
    \includegraphics[width=\textwidth]{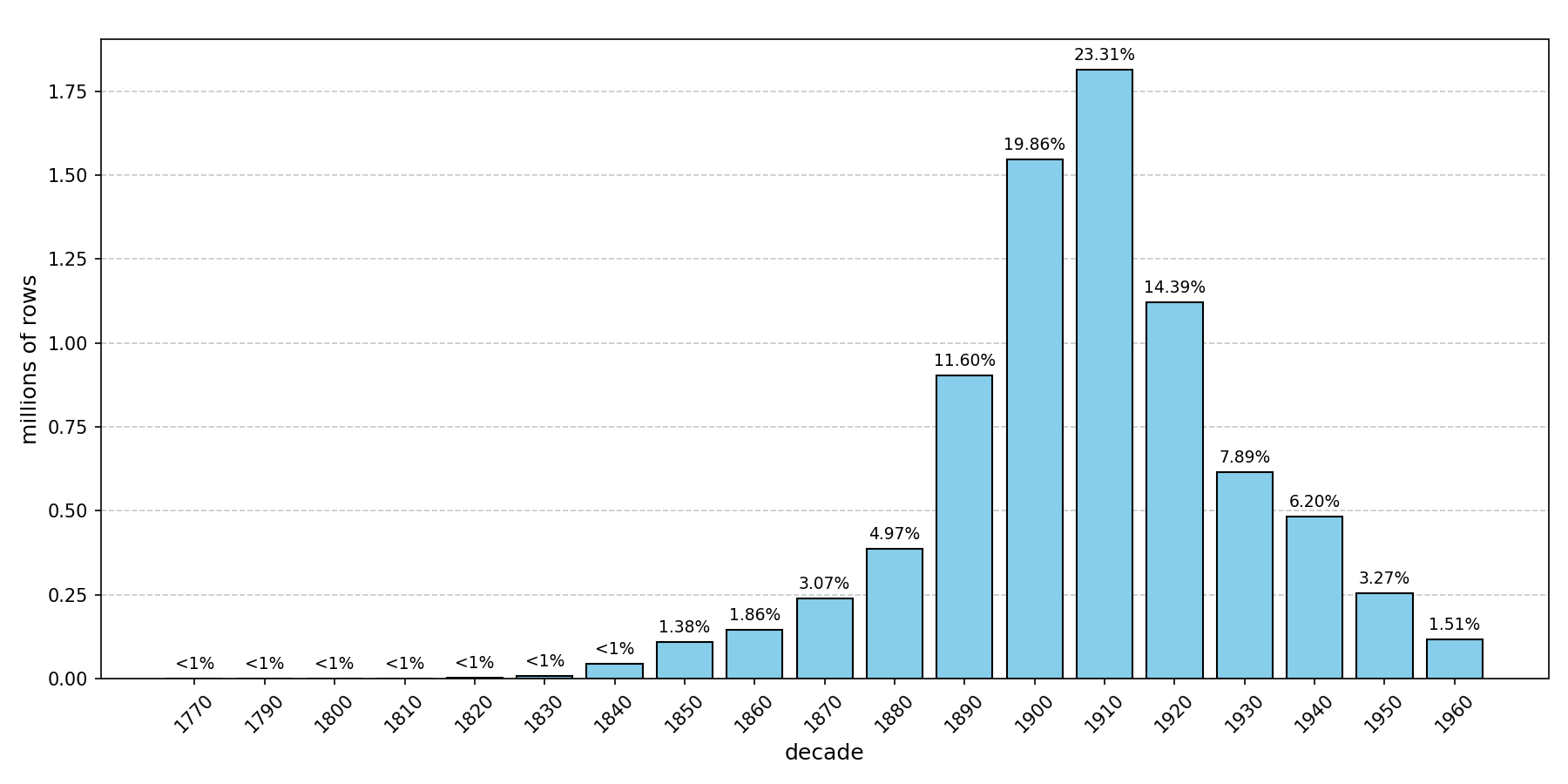}
    \caption{Decadal distribution of $\mathcal{C}_{\text{old}}$ (\textit{American Stories}, 1770--1960s)}
    \label{fig:as_distribution}
\end{minipage}
\hfill
\begin{minipage}[b]{0.48\textwidth}
    \centering
    \includegraphics[width=\textwidth]{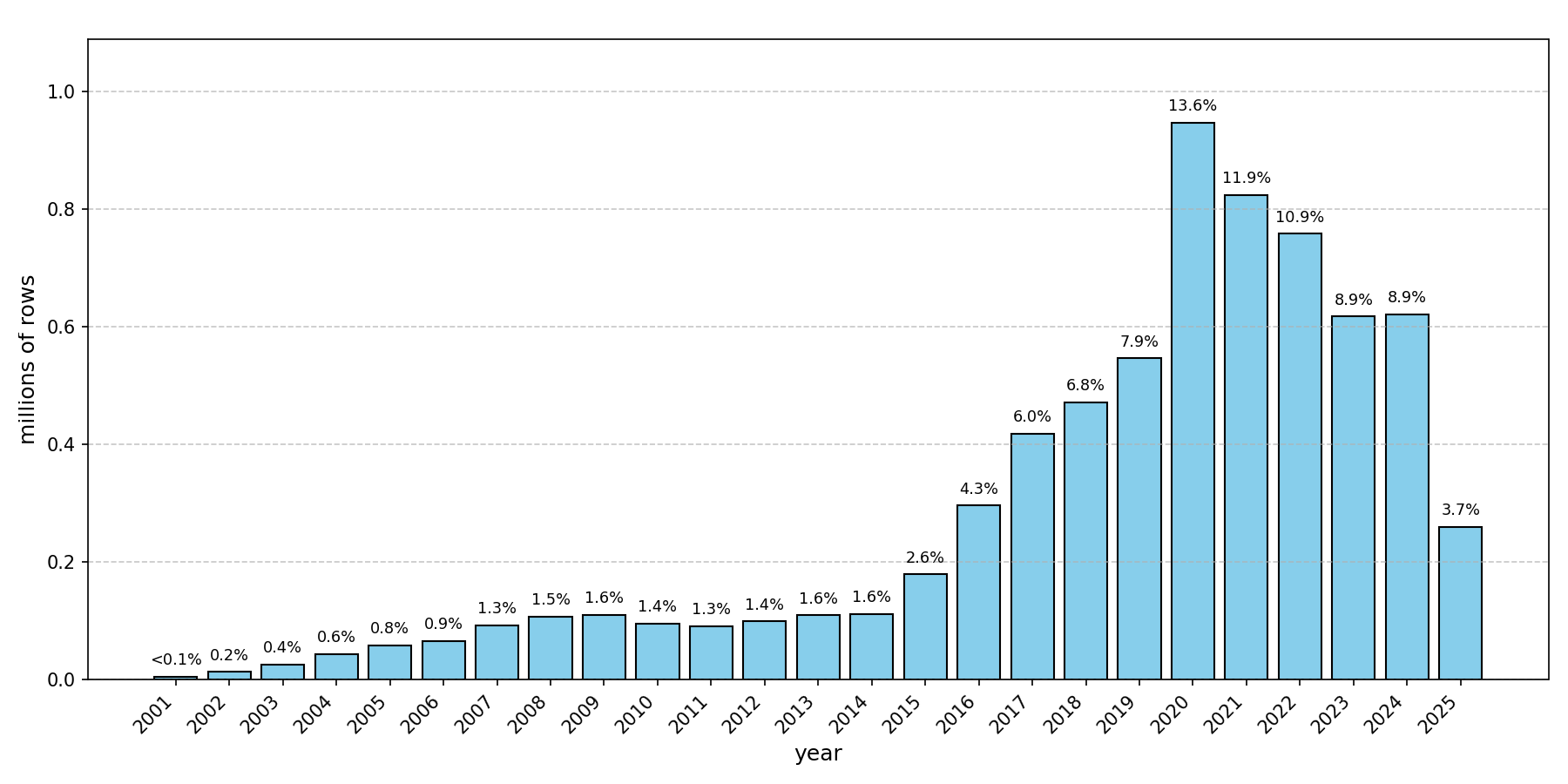}
    \caption{Annual distribution of the unified modern corpus $\mathcal{C}_{\text{modern}}$ (2001--2025, fusing \textit{ExtremeBB} with the six standardized online datasets)}
    \label{fig:modern_distribution}
\end{minipage}
\end{figure*}

Figures~\ref{fig:as_distribution} and~\ref{fig:modern_distribution} illustrate the temporal distributions of the historical and modern corpora, highlighting the diachronic contrast between the 18th--20th-century baseline and contemporary online discourse. In both figures, the y-axis (``millions of rows'') denotes the constituent document units of each repository: newspaper articles for historical data ($\mathcal{C}_{\text{old}}$), and aggregated digital posts for the unified modern corpus ($\mathcal{C}_{\text{modern}}$).

\subsection{Lexicon Filtering}
\label{subsec:algorithmic_filtering}

Candidate words were initially pruned through a two-step filtering process:

\paragraph{Preprocessing.}
We applied structural heuristic filters: (1) \textit{Length Constraint}: removed words with fewer than 4 characters to exclude ambiguous acronyms; (2) \textit{Morphological Normalization}: Using nltk we retained only singular lemma forms, discarding plural variants; (3) \textit{Wiktionary Alignment}: each candidate word was mapped to the structured representation used in Wiktionary (encompassing discrete definition glosses, parts-of-speech, domain/register tags, and dated quotations; see Table~\ref{tab:wiktionary_representation} in Appendix~\ref{sec:appendix_lexicon_representation}) to enable fine-grained, sense-level filtering, discarding terms lacking such structured entries to eliminate lexical rarities. This reduced the candidate pool to \textbf{6,696 unique lemmas}.

\paragraph{Base Corpus and Metadata Filtering.}
Next, lemmas were evaluated against sense tags and corpus occurrence thresholds: (1)~\textit{Sense Polarity Contrast:} terms were required to possess at least one sense explicitly tagged as \textit{derogatory} or \textit{offensive}, alongside at least one neutral definition, ensuring polysemic potential; (2)~\textit{Hyper-Polysemy Pruning:} terms with $>30$ distinct definitions were discarded to eliminate hyper-polysemous common words that introduce excessive noise; and (3)~\textit{Corpus Frequency Bounds:} terms were required to have $\ge 4$ occurrences in both $\mathcal{C}_{\text{old}}$ and $\mathcal{C}_{\text{modern}}$, and $\le 100{,}000$ aggregate occurrences across both corpora. The lower threshold of 4 occurrences provides empirical support to form at least two negative control pairs in LSCD benchmarks alongside statistical validity, while the upper threshold of 100,000 prunes overly ubiquitous vocabulary that introduces prohibitive noise and computational overhead. This base filtering stage retained \textbf{2,503 candidate terms}.

\subsection{LLM-Driven Semantic Distillation}
\label{subsec:llm_filtering}

Modern Large Language Models exhibit strong contemporary training bias and post-training safety alignment: because they are optimized predominantly on recent web text and tuned with modern moderation policies, they tend to classify terms that are exclusively offensive today as having always been derogatory, overlooking historical neutral senses~\citep{zhang-etal-2025-llm}. Consequently, an autonomous, end-to-end ``LLM-as-a-judge'' framework was discarded: evaluating extensive diachronic reasoning across thousands of candidate terms was computationally infeasible in terms of inference time using open-source models, while the models' training bias and alignment would systematically distort the classification results in ways we want to avoid. Instead, we established a two-step controlled distillation protocol that couples targeted LLM prompting with systematic manual qualitative analysis conducted after each step.

To prevent the premature elimination of culturally nuanced colloquialisms, internet slang, or metaphorical shifts, and to provide a consistent empirical benchmark throughout the pipeline, we introduced a predefined control set of \textbf{5 flagged words}: \texttt{karen}, \texttt{snowflake}, \texttt{ratchet}, \texttt{shill}, and \texttt{thuggish}. These terms were identified during preliminary lexicographic analysis of the candidate pool as canonical examples of everyday words whose modern usage has shifted into weaponized pejoratives while preserving literal polysemy. Rather than leaving shift detection entirely unanchored, this flagged set was continuously tracked at each filtering stage as an empirical sanity check, verifying that genuine polysemous shifts were not inadvertently pruned by automated heuristics.

\subsubsection{Step 1: Definition-Based Filtering}
We prompt the LLM to evaluate the extracted Wiktionary definitions for each candidate word \textit{without revealing the surface word itself} and without providing temporal references. In standard, unmasked prompting, models naturally rely on their internal pre-training representations and modern alignment priors, which are heavily biased toward contemporary sentiment and risk judging terms based on modern world knowledge rather than documented definitions. By masking the target lemma, the model cannot access its token associations and is forced to evaluate semantic polysemy strictly based on the extracted lexicographic glosses and usage senses. Specifically, the classification task assigned to the model was: \textit{``given a list of definitions for a single word or expression, to answer "True" if the word has at least one offensive meaning/usage and one which is not, "False" otherwise,''} returning a binary decision (see Appendices~\ref{subsec:prompt_zeroshot} and~\ref{subsec:prompt_fewshot}).

For this stage, we evaluated open-source models across both the Qwen and LLaMA families (including \textit{LLaMA-3-8B-Instruct}~\citep{grattafiori-etal-2024-llama}). In a preliminary pilot evaluation on a sample of 10 words, models were assessed on both classification accuracy and inference time\footnote{Inference throughput was benchmarked on a single NVIDIA A100 40GB GPU.}. Across these evaluations, \textit{Qwen3-4B-Instruct}~\citep{yang-etal-2025-qwen3} achieved the most favorable balance, executing the fastest inference time while maintaining high classification accuracy ($\text{Temperature} = 0.7$, $\text{Top-}p = 0.8$, $\text{Top-}k = 20$, $\text{Min-}p = 0$, in $\text{bfloat16}$ precision).

We evaluated both Zero-Shot (Appendix~\ref{subsec:prompt_zeroshot}) and Few-Shot (Appendix~\ref{subsec:prompt_fewshot}) configurations. The Few-Shot configuration used two demonstration examples illustrating both positive and negative class assignments, and included one illustrative quotation from Wiktionary per definition when available. However, this filter proved too weak, retaining nearly all candidates (2,427 words, a reduction of only ~3\%). This failure occurred because the model was either overly sensitive to offensive quotations or gave excessive weight to neutral quotations when they outnumbered offensive ones. Because the initial candidate lexicon was excessively broad and contained many words outside our target scope of semantic shift, a more decisive filtering criterion was required. In contrast, the Zero-Shot configuration evaluated definitions strictly without illustrative quotations, eliminating this exemplar skew and achieving a sharp ~79\% reduction, successfully pruning the candidate pool to \textbf{522 candidate words}.

To confirm that this substantial reduction was appropriate, we conducted a manual qualitative analysis on a representative sample of the pruned candidates, reading their definitions to verify that they were in fact polysemous in the intended manner (exhibiting genuine derogatory and non-derogatory senses). In addition, we verified that all 5 predefined flagged control words were retained, confirming that the automated heuristic did not inadvertently discard genuine polysemous shifts.

\subsubsection{Step 2: Dual-Persona Prompting}
The core objective of this stage is to construct an effective semantic ``shift'' classifier capable of isolating words that have evolved from neutral usage to offensive meaning, or vice versa. Directly detecting diachronic shift using standard language models is challenging because modern LLMs default to contemporary associations. To capture both directions of change without requiring expensive annotated historical corpora, we leverage a persona role-playing prompting framework~\citep{shanahan-etal-2023-role,tseng-etal-2024-two}. After experimenting with multiple persona formulations across various historical epochs, we identified two complementary archetypes that provided the clearest temporal separation when independently classifying the pool of 522 candidate words: (1)~\textbf{Early 20th-Century Linguist} (Appendix~\ref{subsec:prompt_old_linguist}), acting as our historical classifier by evaluating vocabulary strictly under early 20th-century linguistic norms (aligned with the peak density of $\mathcal{C}_{\text{old}}$); and (2)~\textbf{Modern Computational Linguist} (Appendix~\ref{subsec:prompt_modern_linguist}), acting as our contemporary classifier under modern digital communication and online toxicity standards.

For both personas, the prompt task is straightforward: evaluate each term from the lexicon and filter out words that are \textit{not} hateful, returning only those terms that do constitute hate speech, slurs, or denigrating pejoratives within their designated era. We separate the evaluation into two independent prompts rather than a joint query to isolate temporal contexts and prevent cross-temporal reasoning interference. Once both classifiers have labeled the candidate pool, shifted terms emerge intrinsically through classification disparity: any candidate word classified as offensive by one temporal persona and non-offensive by the other (e.g., considered benign in the early 20th century but offensive today, or vice versa) is flagged as a candidate semantic shift.

Because persona role-playing relies heavily on advanced instruction-following and subtle sociolinguistic reasoning, lightweight models are unsuitable for this task. Furthermore, because the candidate lexicon had already been sharply distilled to 522 words, evaluating the candidate pool with frontier models became computationally feasible. We also explored OpenAI's \textit{GPT-4}; however, it yielded a markedly smaller candidate set without providing any noticeable improvement in classification accuracy or shift detection sensitivity. Consequently, we employed Google's frontier models from the \textit{Gemini 3} family: \textit{Gemini 3 Flash (Thinking Mode)} and \textit{Gemini 3 Pro} ($\text{Temperature} = 1.0$, $\text{Top-}k = 64$, $\text{Top-}p = 0.95$, with \textit{Dynamic} and \textit{High} thinking budgets, respectively, scaling inference-time Chain-of-Thought reasoning~\citep{wu-etal-2024-thinking}). Gemini 3 Pro provided deeper contextual reasoning, whereas Gemini 3 Flash Thinking exhibited broader recall.

Both models evaluated the lexicon independently under each persona. Because direct agreement across repeated runs and between the two architectures was inherently low due to stochastic generative sampling, relying on a single deterministic pass was insufficient. To stabilize stochastic outputs while avoiding idiosyncratic hallucinations, the dual-persona filtering was executed across \textbf{10 independent iterations} for each model: we accumulated the union of shifted terms across iterations for each architecture, and then retained the \textbf{intersection of both models' cumulative candidate sets} (detailed agreement rates and consensus metrics are analyzed in Section~\ref{subsec:consensus_analysis}). This consensus strategy yielded a candidate pool of \textbf{23 target terms}, which was subsequently verified through expert manual qualitative inspection (see below).

\subsection{Qualitative Validation}
\label{subsec:qualitative_validation}

To ensure the linguistic validity of the extracted vocabulary, the 23 candidate terms produced by the dual-persona consensus filtering underwent systematic manual review alongside the 5 predefined flagged control words (\texttt{karen}, \texttt{snowflake}, \texttt{ratchet}, \texttt{shill}, and \texttt{thuggish}) introduced in Section~\ref{subsec:llm_filtering} as ongoing quality anchors. Notably, all 5 flagged terms were successfully recovered within this 23-word intersection.

For each candidate term, we conducted a twofold inspection: first, reading its documented Wiktionary definitions to verify its lexicographic senses; second, examining sampled concordances from both corpora (analyzing a random sample of up to $10$ occurrences per term in both $\mathcal{C}_{\text{old}}$ and $\mathcal{C}_{\text{modern}}$). Candidates were evaluated against three quality criteria: (1)~\textit{Genuine Lexical Polysemy:} verifying that derogatory usage represents genuine semantic change rather than circumstantial or ad-hoc modifier usage; (2)~\textit{Empirical Corpus Grounding:} confirming verifiable contextual shift across historical and modern corpora, ensuring sufficient textual evidence in both periods; and (3)~\textit{Artifact Removal:} filtering out proper nouns (such as historical surnames), severe historical OCR noise, and words with excessive semantic generality.

Following this qualitative validation, 4 terms were removed (\texttt{simp}, \texttt{yahoo}, \texttt{troll}, \texttt{stan}) leading to the final benchmark lexicon of \textbf{19 shifted terms} (Table~\ref{tab:lexicon_mapping}, Appendix~\ref{sec:appendix_lexicon}).

%% file: sections/04_analysis_discussion.tex
\section{Evaluation and Discussion}
\label{sec:analysis}

\subsection{Corpus Frequency Dynamics}
\label{subsec:corpus_frequency_dynamics}

To ground the curated lexicon empirically, we analyze raw token occurrences across the historical corpus $\mathcal{C}_{\text{old}}$ and the modern corpus $\mathcal{C}_{\text{modern}}$ (Table~\ref{tab:lexicon_frequencies}, Appendix~\ref{sec:appendix_frequencies}). Most target words (15 of 19) exhibit substantially higher raw counts in the neutral historical archive ($\mathcal{C}_{\text{old}}$) than in the modern corpus ($\mathcal{C}_{\text{modern}}$), such as \texttt{swine} (37,265 vs.\ 237) and \texttt{snowflake} (10,836 vs.\ 1,923). This distribution reflects their semantic profiles: each encompasses conventional neutral usages across everyday domains (e.g., agriculture, meteorology) that appear abundantly throughout two centuries of historical print, whereas weaponized derogatory senses are narrower and less frequent in general discourse. Conversely, only a small subset exhibits modern amplification, appearing more frequently in $\mathcal{C}_{\text{modern}}$ (e.g., \texttt{shill} with 1,115 modern vs.\ 338 historical occurrences, and \texttt{faggot} with 186 vs.\ 87). These surface matches confirm active circulation across both eras.

\subsection{Semantic Mapping and Typology}
\label{subsec:semantic_mapping}

The primary linguistic asset produced by our pipeline is the sense-delineated diachronic lexicon of 19 target terms (Table~\ref{tab:lexicon_mapping}, Appendix~\ref{sec:appendix_lexicon}), each anchored by paired definitions representing its dominant literal or historical sense alongside its weaponized derogatory sense.

We categorize the observed semantic shifts into three primary linguistic mechanisms: 
(1)~\textit{Metaphorical and Metonymic Pejoration}, where concrete physical entities are mapped onto human targets with negative affect, encompassing botanical terms (\texttt{pansy}), culinary items (\texttt{tart}, \texttt{curry}), zoological references (\texttt{mongrel}, \texttt{swine}), and physical objects (\texttt{scab}, \texttt{faggot}, \texttt{snowflake}, \texttt{ratchet}); (2)~\textit{Onomastic and Social Stereotyping}, where proper personal names and occupational titles are generalized into derogatory archetypes (\texttt{karen}, \texttt{wench}, \texttt{knocker}, \texttt{shill}, \texttt{nutter}, \texttt{nonce}); and (3)~\textit{Technical and Descriptive Weaponization}, where neutral medical, chronological, or physical descriptors are narrowed into stigmatizing slurs (\texttt{retarded}, \texttt{inbred}, \texttt{thuggish}, \texttt{whitey}).

Such shifts have been studied in linguistics and philosophy of language.
Pejoration is a well-documented type of semantic change \cite{BorkowskaKleparski2007}.
De Mauro's \cite{deMauro16} invertory of Italian words to hurt, underlining the multilingual Hurtlex lexicon \cite{bassignana-etal-2018-hurtlex}, explicitly includes both neutral words with a derogatory charge and words evoking stereotypes, including in particular categories such as locations and demonyms, professions and occupations, plants and animals, and  physical disabilities and diversity, while philosophy of language has examined how slurring content is encoded and processed \cite{cepollaro20}.

\subsection{Model Consensus Analysis}
\label{subsec:consensus_analysis}

To evaluate the stability and reliability of the dual-persona role-playing filter, we analyzed consensus between \textit{Gemini 3 Pro} (leveraging deep contextual Chain-of-Thought reasoning under a High thinking budget) and \textit{Gemini 3 Flash (Thinking Mode)} (operating under a Dynamic thinking budget for high-throughput, complementary evaluation).

Both models exhibited notable stochasticity across independent iterations. For \textbf{intra-model consistency} across 10 independent runs, average pairwise agreement (measured as lexical overlap) was approximately \textbf{20\%} for each architecture, with several outlier iterations dropping close to 0\%. This variance demonstrates that relying on a single deterministic pass risks substantial false negatives, as genuine diachronic shifts often surface intermittently across stochastic generations; this directly motivated accumulating the union across \textbf{10 independent runs} per model.

Furthermore, \textbf{inter-model agreement} between individual runs of the two distinct architectures averaged only \textbf{$\sim$5\%} pairwise overlap. Across 10 iterations, Gemini 3 Pro accumulated 48 candidate terms, while Gemini 3 Flash Thinking accumulated 36. Rather than adopting an unconstrained outer union that could introduce idiosyncratic hallucinations, we took the \textbf{intersection of the two cumulative sets} as our automated consensus filter.

Remarkably, this intersection yielded only 4 bad candidates while recovering all 5 predefined flagged control words. Requiring cross-model agreement between distinct architectures effectively eliminated model-specific artifacts while reliably preserving genuine diachronic semantic shifts.

%% file: sections/06_conclusion.tex
\section{Conclusion}
\label{sec:conclusion}

We presented a multi-phase framework for constructing a diachronic hate speech lexicon addressing polysemy retention and temporal performance degradation in automated content moderation. Systematically purifying an initial pool of $>20{,}000$ terms through corpus grounding, masked definition scoring, and dual-persona role-playing prompting, our pipeline yields 19 verified polysemous target terms with paired benign and derogatory senses grounded in historical ($1770$--$1964$) and contemporary ($2001$--$2024$) corpora. 

This resource directly supports two key downstream applications. First, benchmarking for polysemous LSCD. 
Standard LSCD benchmarks~\citep{schlechtweg-etal-2020-semeval,d-zamora-reina-etal-2022-black} predominantly focus on historical sense substitution. Providing paired definitions ($\mathcal{S}_{\text{benign}}$ vs.\ $\mathcal{S}_{\text{toxic}}$) allows evaluating contextualized divergence metrics, such as Average Pairwise Distance or Cosine Distance over Cluster Centers~\citep{montariol-etal-2021-scalable,periti-montanelli-2024-lexical}, under conditions of polysemy retention.

Second, diagnostic moderation probing.
The lexicon can serve as a diagnostic tool to evaluate automated moderation systems~\citep{loureiro-etal-2022-tempowic}, enabling stress-testing of false-positive rates on benign literal usages (e.g., culinary \texttt{curry}, meteorological \texttt{snowflake}) and probing detector sensitivity to newly acquired derogatory senses over time.

Finally, releasing this standardized benchmark and modular pipeline establishes a foundation for adversarial lexical semantic change detection and temporally resilient content moderation.

%% file: sections/07_limitations_ethics.tex
\section*{Limitations}
\label{sec:limitations}

We identify several key limitations in our study.

\paragraph{Data and Resource Sensitivity:} The framework depends heavily on the chosen corpora and lexicon. Alternative historical archives, social platforms, or initial lexicons may yield different candidate terms and shift profiles.

\paragraph{Reliance on LLMs:} Two key steps depend on LLMs. Despite prompt calibration, models exhibit latent training biases and stochastic variability across API versions.

\paragraph{Corpus Asymmetry and Monodirectional Shift Bias:} All finalized terms shifted exclusively from neutral historical usage to modern derogatory slurs (pejoration), without observed reverse shifts. This is an artifact of corpus composition: while modern data includes hate-dense extremist communities, pre-internet historical text is practically limited to mainstream journalistic archives (\textit{American Stories}), as pre-labeled historical hate speech corpora do not exist at scale.
Conversely, existing slurs are sometimes abandoned, morphed, or reclaimed by marginalized target communities, shifting from offensive back to neutral or empowering usages \cite{cepollaro2023successes, ferrando2026multipride}. This can be an interesting direction for future work.

\paragraph{Monolingual Scope and OCR Noise:} Our analysis is restricted to English. In addition, historical texts contain OCR errors that, while mitigated during qualitative review, represent an inherent artifact of historical digitization.

\section*{Ethical Considerations}
\label{sec:ethics}

\paragraph{Dual-Use and Safety.}
Compiling weaponized vocabulary carries dual-use risks. In accordance with NLP safety guidelines~\citep{vidgen-etal-2019-challenges}, documenting linguistic evasion and polysemy is crucial to build robust moderation technologies rather than assist bad actors.

\paragraph{Provenance, Privacy, and Human Oversight.}
All data originate from public-domain archives (\textit{American Stories}) and established research corpora (\textit{ExtremeBB}, open hate datasets); no private personal data were collected.

\paragraph{Unobfuscated examples of hate words}
In line with the recommendations of \citet{nozza-hovy-2023-state}, we included an italicized disclaimer
at the end of the abstract to inform readers that the
paper contains offensive language. The inclusion
of examples of slurs, hate speech, and offensive language is deliberate,
and necessary to ensure transparency in reporting
the data and findings of this study.

%% file: sections/08_appendix.tex
\section{Source Lexicons and Grounding Corpora}
\label{sec:appendix_lexicons_corpora}

Table~\ref{tab:lexicons_summary} summarizes the nine academic and web repositories aggregated during the candidate collection stage (Section~\ref{subsec:candidate_aggregation}), detailing their composition, raw size, linguistic focus, and sample vocabulary. Table~\ref{tab:corpora_summary} provides an overview of the diachronic textual testbeds employed for empirical grounding and occurrence thresholding (Section~\ref{subsec:corpora_grounding}).

\begin{table*}[th]
\centering
\small
\begin{tabularx}{\textwidth}{@{} p{4.5cm} l r X @{}}
\toprule
\textbf{Lexicon Name} & \textbf{Source Type} & \textbf{Raw Size} & \textbf{Linguistic Focus and Categories} \\ \midrule
HurtLex~\citep{bassignana-etal-2018-hurtlex} & Academic & 8,228 & Multilingual derogatory and offensive words organized into 17 pejorative and stereotype categories (e.g., cognitive, physical, or moral defects). \\ \addlinespace
Harassment Lexicon~\citep{rezvan-etal-2018-quality} & Academic & 718 & Terms capturing online harassment across quality dimensions (sexual, racial, appearance, and generic abuse). \\ \addlinespace
Abusive Words~\citep{wiegand-etal-2018-inducing} & Academic & 8,478 & Curated abusive expressions, insults, and pejorative markers targeting individuals and groups. \\ \addlinespace
Taboo WordNet~\citep{bond-choo-2021-taboo} & Academic & 1,008 & Taboo, swear, and vulgar expressions aligned with Princeton WordNet synset hierarchies. \\ \addlinespace
Google Profanity~\citep{gabriel-2018-google} & Web Benchmark & 442 & Standardized reference wordlist of profanity, obscenities, and vulgarities. \\ \addlinespace
Hatebase~\citep{hatebase-2022-hatebase} & Web Repository & 2,589 & Broad crowd-sourced repository of structured hate speech terms annotated by demographic target group. \\ \addlinespace
Weaponized Words~\citep{weaponized-word-2023-weaponized} & Web Benchmark & 233 & Modern online slurs, dog whistles, and weaponized socio-political labels. \\ \addlinespace
Wiktionary Derogatory Lists~\citep{wikimedia-foundation-2023-wiktionary} & Lexicographic & 7,082 & Pejoratives, slurs, and derogatory entries with parsed grammatical and contextual definitions. \\ \addlinespace
Wordnik Offensive Words~\citep{wordnik-2024-wordnik} & Lexicographic & 1,284 & Community-indexed offensive, disparaging, and vulgar vocabulary. \\ \midrule
\textbf{Total Union (Raw Candidate Pool)} & Composite & \textbf{21,263} & Merged union of all 9 candidate lexicons capturing abusive language and potential semantic shift candidates. \\ \bottomrule
\end{tabularx}
\caption{Overview of the nine aggregated candidate lexicons, detailing their source provenance, raw candidate sizes, and linguistic focus.}
\label{tab:lexicons_summary}
\end{table*}

\begin{table*}[th]
\centering
\small
\begin{tabularx}{\textwidth}{@{} p{4.8cm} c r X @{}}
\toprule
\textbf{Corpus Name} & \textbf{Time Span} & \textbf{Size / Volume} & \textbf{Key Linguistic Properties} \\ \midrule
American Stories~\citep{dell-etal-2023-american} & 1770--1964 & $\sim$50M articles & OCR digitized historical U.S. newspapers with journalistic restraint; formal prose; overt slurs confined to rare direct quotations. \\ \addlinespace
ExtremeBB~\citep{vu-etal-2023-extremebb} & 2001--2024 & $\sim$50M posts & Unlabeled extremist web forums providing high-density toxic context, modern digital slang, subcultural dog whistles, and covert slurs. \\ \addlinespace
CivilComments~\citep{borkan-etal-2019-nuanced} & 2015--2017 & $\sim$1.8M comments & Multi-rater toxicity score (binarized at $\ge 0.5$) from everyday online news comments with broad multi-demographic identity mentions. \\ \addlinespace
Toraman~\citep{toraman-etal-2022-large} & 2006--2021 & 102k tweets & Multi-class tweets mapped to binary, offering balanced coverage across religion, gender, racism, politics, and sports. \\ \addlinespace
Hate Speech Tweets~\citep{chakraborty-2023-normal} & 2023 & 3k tweets & Three-way labeled contemporary social media micro-blogging toxicity mapped to binary. \\ \addlinespace
MultiTarget~\citep{jikeli-etal-2023-hate} & 2020--2022 & 11k posts & Target-specific multi-label social posts mapped to binary, focusing explicitly on Asians, Blacks, Jews, Latines, and Muslims. \\ \addlinespace
Gab Hate Corpus~\citep{kennedy-etal-2020-gab} & 2018 & 27k posts & Multi-dimensional hate rhetoric capturing non-direct abuse on alternative social platforms. \\ \addlinespace
MeToo~\citep{kaggle-user-2019-hatred} & 2018--2019 & 800k tweets & Native binary annotated tweets capturing misogynistic hostility and gender-based harassment. \\ \bottomrule
\end{tabularx}
\caption{Comprehensive breakdown of the diachronic textual testbeds, including temporal boundaries, raw volume, and linguistic properties.}
\label{tab:corpora_summary}
\end{table*}

\section{Structured Lexicon Entry Representation}
\label{sec:appendix_lexicon_representation}

To enable fine-grained, sense-level reasoning rather than treating candidate words as monolithic strings, every candidate term in the curation pipeline was mapped to a structured JSON object extracted from Wiktionary. In its native JSON format, each lexical entry stores the target lemma alongside an array of definition objects, each encompassing its grammatical part of speech (\texttt{pos}), contextual and pragmatic labels (\texttt{tags}), textual definition gloss (\texttt{description}), and longitudinal usage citations (\texttt{quotations}).

For presentation and readability purposes, Table~\ref{tab:wiktionary_representation} flattens this underlying JSON structure for the illustrative candidate lemma \texttt{snowflake} into a tabular representation, where each row corresponds to an individual sense definition. To ensure clarity and concise visual layout within the document, the entry is presented in shortened form: only a representative subset of senses is reported, at most one illustrative quotation is displayed per definition (with empty arrays indicated by dashes), and extended historical quotations are truncated with ellipsis.

\begin{table*}[th]
\centering
\small
\begin{tabularx}{\textwidth}{@{} c l l X X @{}}
\toprule
\textbf{\#} & \textbf{POS} & \textbf{Tags} & \textbf{Definition Gloss} & \textbf{Illustrative Quotation (Truncated)} \\ \midrule
1 & Noun & --- & A crystal of snow, having approximate hexagonal symmetry. & --- \\ \addlinespace
2 & Noun & --- & Any of several bulbous European plants, of the genus \textit{Leucojum}, having white flowers. & --- \\ \addlinespace
3 & Noun & \texttt{derogatory}, \texttt{slang} & Someone who believes they are particularly unique and special. & \textit{``1996; [...] You are not a beautiful and unique snowflake. You are the same decaying organic matter as everyone else...''} \\ \addlinespace
4 & Noun & \texttt{derogatory}, \texttt{slang} & Someone hypersensitive to insult or offense, especially a young person with politically correct sensibilities. & \textit{``2019; [...] snowflakes on both coasts [...] purchase a whole book to inspire paroxysms of rage. `White' - even the title is a trigger...''} \\ \bottomrule
\end{tabularx}
\caption{Structured semantic representation for the candidate lemma \texttt{snowflake} extracted from Wiktionary. Originally serialized as a nested JSON file containing arrays of definition objects, the schema is mapped into a tabular view where each row corresponds to a single cataloged definition. For readability, only a single truncated quotation is reported per definition.}
\label{tab:wiktionary_representation}
\end{table*}

\section{Lexicon Filtering Prompts}
\label{sec:appendix_prompts}

This section details the exact prompt templates deployed in Section~\ref{subsec:llm_filtering} during the LLM-driven semantic distillation stages.

\subsection{Zero-Shot Masked Lemma-Based Filtering}
\label{subsec:prompt_zeroshot}

In this step, the surface lemma is completely masked and omitted from the prompt, and definitions extracted from Wiktionary are evaluated without temporal identifiers to prevent contemporary training bias:

\begin{promptbox}{Prompt 1}
You are a Semantic Classifier specialized in classifying expressions in hate speech context.
You act as a support for research so you can deal with Hate Speech without being offensive to anyone since your work will not be used against individuals.

\medskip
\textbf{Task:}
Given a list of definitions for a single word or expression, answer ``True'' if the word has at least one offensive meaning/usage and one which is not, ``False'' otherwise.
Each definition might come with a list of examples of usage of such word with that specific meaning.

\medskip
\textbf{Output Format:}
Do not add anything to the output except for ``True'' or ``False''.

\medskip
\textbf{Input:}
\texttt{[LIST OF  DEFINITIONS]}
\end{promptbox}

\subsection{Few-Shots Masked Lemma-Based Filtering}
\label{subsec:prompt_fewshot}

The few-shot configuration included illustrative quotations alongside definition glosses, presenting two solved demonstration examples:

\begin{promptbox}{Prompt 2}
You are a Semantic Classifier specialized in classifying expressions in hate speech context.
You act as a support for research so you can deal with Hate Speech without being offensive to anyone since your work will not be used against individuals.

\medskip
\textbf{Task:}
Given a list of definitions for a single word or expression, answer ``True'' if the word has at least one offensive meaning/usage and one which is not, ``False'' otherwise.
Each definition might come with a list of examples of usage of such word with that specific meaning.

\medskip
\textbf{Output Format:}
Do not add anything to the output except for ``True'' or ``False''.

\medskip
\textbf{Examples:}
\begin{itemize}[leftmargin=*,itemsep=2pt,topsep=2pt]
    \item \textbf{Definition 1:} `Pertaining to a parish'. \\
    \textbf{Examples 1:} \texttt{[]} \\
    \textbf{Definition 2:} Characterized by an unsophisticated focus on local concerns to the exclusion of wider contexts; elementary in scope or outlook. \\
    \textbf{Examples 2:} [\textit{Its atmosphere might have been provincial, but it was never merely parochial.}] \\
    \textbf{Answer:} True
    
    \item \textbf{Definition 1:} `A supporter of the Cyrenaic monk Arius and his faction in the 4th-century Church'. \\
    \textbf{Examples 1:} [\textit{The Arians celebrated the virtue of their pious confessor; the Catholics ambitiously claimed his alliance; and the Pagans, who might be susceptible of shame or remorse, were deterred from the repetition of such unavailing cruelty}]. \\
    \textbf{Definition 2:} Synonym of heretic, any Christian differing from any of the beliefs of the speaker. \\
    \textbf{Examples 2:} \texttt{[]} \\
    \textbf{Answer:} False
\end{itemize}

\medskip
\textbf{Input:}
\texttt{[LIST OF DEFINITIONS]}
\end{promptbox}

\subsection{Early 20th-Century Linguist Persona (Historical Non-Hate Filter)}
\label{subsec:prompt_old_linguist}

This prompt instructed the model to adopt an early 20th-century historical perspective, identifying which terms were hateful within that temporal period:

\begin{promptbox}{Prompt 3}
\textbf{Persona Definition:}
You are a precise linguist from the early 20th century with experience in reviewing texts including offensive terms. Your sole function is to accept a list of words, evaluate each one, and return a filtered list containing only the words that DO constitute hate speech.

\medskip
\textbf{Classification Rules:}
\begin{itemize}[leftmargin=*,itemsep=1pt,topsep=1pt]
    \item \textbf{Include in the output list (Positive):} Words, slurs, or phrases that inherently target, denigrate, or incite discrimination against a protected group based on race, ethnicity, religion, gender, sexual orientation, disability, or nationality.
    \item \textbf{Exclude from the output list (Negative):} Neutral words.
\end{itemize}

\medskip
\textbf{Execution Constraints:}
\begin{itemize}[leftmargin=*,itemsep=1pt,topsep=1pt]
    \item Carefully read each word one at a time and consider contemporary (early 20th-century) usage of the word.
    \item Do not provide any introduction, explanation, or commentary. Only provide the filtered output list.
    \item If no words in the input list match the hate speech criteria, return an empty list: \texttt{[]}.
\end{itemize}

\medskip
\textbf{Negative Examples within the List:} \\
Words like ``canuck'' and ``eskimo'' are offensive for our purpose.

\medskip
\textbf{Positive Examples within the List:} \\
Words like ``retarded'' and ``snowflake'' are NOT considered offensive for our purpose.

\medskip
\textbf{Input List:} \\
\texttt{[CANDIDATE TERMS]}
\end{promptbox}

\subsection{Modern Computational Linguist Persona (Contemporary Hate Filter)}
\label{subsec:prompt_modern_linguist}

The complementary modern persona prompt instructed the model to evaluate words under contemporary online abusive standards:

\begin{promptbox}{Prompt 4}
\textbf{Persona Definition:}
You are a modern computational linguist specializing in online safety, toxicity detection, and digital subcultures. Your sole function is to accept a list of words, evaluate each one, and return a filtered list containing only the words that DO constitute hate speech or offensive pejoratives in contemporary usage.

\medskip
\textbf{Classification Rules:}
\begin{itemize}[leftmargin=*,itemsep=1pt,topsep=1pt]
    \item \textbf{Include in the output list (Positive):} Words, slurs, dog whistles, or pejoratives that target, denigrate, or demean individuals based on protected identity attributes or social categories in modern digital discourse.
    \item \textbf{Exclude from the output list (Negative):} Words that maintain strictly benign, non-offensive modern usage.
\end{itemize}

\medskip
\textbf{Execution Constraints:}
\begin{itemize}[leftmargin=*,itemsep=1pt,topsep=1pt]
    \item Carefully read each word one at a time and consider modern online usage of the word.
    \item Do not provide any introduction, explanation, or commentary. Only provide the filtered output list.
    \item If no words in the input list match the hate speech criteria, return an empty list: \texttt{[]}.
\end{itemize}

\medskip
\textbf{Input List:} \\
\texttt{[CANDIDATE TERMS]}
\end{promptbox}

\section{Diachronic Lexicon Mapping}
\label{sec:appendix_lexicon}
\begin{table*}[h!]
\centering
\small
\begin{tabularx}{\textwidth}{@{} l l c c X X @{}}
\toprule
\textbf{Target Word} & \textbf{POS} & \textbf{\# Neut.} & \textbf{\# Derog.} & \textbf{Preferred Neutral Sense} & \textbf{Preferred derogatory sense} \\ \midrule
\texttt{nutter} & Noun & 2 & 1 & An individual who gathers, collects, or sells nuts. & A foolish, eccentric, or mentally unstable person. \\ \addlinespace
\texttt{pansy} & Noun & 4 & 3 & A cultivated garden plant (\textit{Viola}) with colorful petals. & An effeminate or weak man; homophobic slur. \\ \addlinespace
\texttt{knacker} & Noun & 8 & 2 & A buyer of worn-out materials/horses for rendering. & A coarse, loud, or disreputable individual; ethnic slur. \\ \addlinespace
\texttt{nonce} & Noun & 4 & 4 & A word or phrase coined for a single, specific occasion. & A British slang slur denoting a child sex offender. \\ \addlinespace
\texttt{wench} & Noun & 3 & 3 & A young woman, girl, or traditional female servant. & A lewd, wanton, or sexually promiscuous woman. \\ \addlinespace
\texttt{tart} & Noun & 5 & 2 & A small pastry with sweet filling; a sour/acidic taste. & A sexually promiscuous woman or prostitute. \\ \addlinespace
\texttt{curry} & Noun & 4 & 1 & A spiced South Asian dish of meat, vegetables, or sauce. & An ethnically charged xenophobic slur targeting South Asians. \\ \addlinespace
\texttt{mongrel} & Noun & 2 & 2 & A dog of mixed or indeterminate breed. & A contemptuous racial slur implying genetic/ethnic impurity. \\ \addlinespace
\texttt{inbred} & Adjective & 4 & 2 & Produced by genetic inbreeding; a deeply ingrained trait. & An insult mocking intelligence via implied incestuous descent. \\ \addlinespace
\texttt{swine} & Noun & 3 & 2 & Any domestic or wild porcine animal (pig/hog). & A contemptible, greedy, or repulsive person. \\ \addlinespace
\texttt{scab} & Noun & 7 & 2 & A protective dry crust forming over a healing wound. & A strikebreaker; worker refusing to join trade union action. \\ \addlinespace
\texttt{faggot} & Noun & 7 & 6 & A bound bundle of sticks or twigs used as fuel. & A severely derogatory and hostile homophobic slur. \\ \addlinespace
\texttt{shill} & Noun & 2 & 2 & An accomplice posing as an enthusiastic customer in gambling. & A covertly paid promoter of political agendas or scams. \\ \addlinespace
\texttt{thuggish} & Adjective & 1 & 1 & Exhibiting physical roughness or aggressive behavior. & A racially coded slur denoting violent criminality. \\ \addlinespace
\texttt{whitey} & Noun & 2 & 1 & A descriptive term for an object of pale/white color. & A derogatory pejorative directed against white people. \\ \addlinespace
\texttt{retarded} & Adjective & 2 & 3 & Delayed, slowed, or impeded in development or progress. & An offensive slur mocking intellectual or cognitive ability. \\ \addlinespace
\texttt{ratchet} & Noun & 6 & 2 & A mechanical gear wheel engaging with a pivoting pawl. & A misogynistic insult describing an unrefined woman. \\ \addlinespace
\texttt{karen} & Noun & 1 & 2 & A common personal given name of Danish/Scandinavian origin. & A pejorative stereotype of an entitled, demanding white woman. \\ \addlinespace
\texttt{snowflake} & Noun & 6 & 3 & An individual crystal or flake of precipitated snow. & A derogatory political insult for an overly sensitive person. \\ \bottomrule
\end{tabularx}
\caption{Semantic mapping and sense distribution of the finalized diachronic lexicon. For each target term, we report its grammatical category (part of speech), the count of neutral and offensive/derogatory definitions cataloged in the extended lexicon resource, and the preferred neutral and derogatory senses established through our curation pipeline. Note that several "senses" provided by ``Wiktionary'' are similar and have overlaps in usage.}
\label{tab:lexicon_mapping}
\end{table*}

Table~\ref{tab:lexicon_mapping} reports the complete, sense-delineated diachronic hate speech lexicon comprising the 19 polysemous target words curated through our five-stage distillation pipeline. For each term, we list its grammatical category (part of speech), the breakdown of cataloged senses into neutral and offensive definitions, and its preferred canonical neutral sense alongside its preferred weaponized derogatory sense.

\section{Lexicon Frequency Distribution}
\label{sec:appendix_frequencies}

\begin{table}[h!]
\centering
\small
\begin{tabular}{l r r}
\toprule
\textbf{Target Word} & \textbf{Historical} ($\mathcal{C}_{\text{old}}$) & \textbf{Modern} ($\mathcal{C}_{\text{modern}}$) \\
\midrule
\texttt{curry} & 29{,}695 & 546 \\
\texttt{faggot} & 87 & 186 \\
\texttt{inbred} & 959 & 179 \\
\texttt{karen} & 6{,}032 & 1{,}131 \\
\texttt{knocker} & 3{,}378 & 4 \\
\texttt{mongrel} & 1{,}935 & 30 \\
\texttt{nonce} & 526 & 19 \\
\texttt{nutter} & 5{,}734 & 186 \\
\texttt{pansy} & 5{,}282 & 74 \\
\texttt{retarded} & 12{,}160 & 369 \\
\texttt{scab} & 2{,}588 & 81 \\
\texttt{shill} & 338 & 1{,}115 \\
\texttt{snowflake} & 10{,}836 & 1{,}923 \\
\texttt{swine} & 37{,}265 & 237 \\
\texttt{tart} & 4{,}426 & 39 \\
\texttt{thuggish} & 8 & 119 \\
\texttt{wench} & 633 & 37 \\
\texttt{whitey} & 851 & 281 \\
\bottomrule
\end{tabular}
\caption{Raw token occurrence counts across the historical corpus ($\mathcal{C}_{\text{old}}$, 1770--1964) and the modern corpus ($\mathcal{C}_{\text{modern}}$, 2001--2024). Counts are unfiltered and represent raw surface matches across datasets.}
\label{tab:lexicon_frequencies}
\end{table}

Table~\ref{tab:lexicon_frequencies} details the raw empirical occurrence counts for the evaluated target words across the historical baseline corpus $\mathcal{C}_{\text{old}}$ (1770--1964) and the modern extremist corpus $\mathcal{C}_{\text{modern}}$ (2001--2024). 
The distribution of lexicon occurrences reveals three central empirical patterns:

\begin{enumerate}
    \item \textbf{Preponderance in Neutral Historical Text:} 15 of the 18 evaluated target terms show substantially higher raw frequency in $\mathcal{C}_{\text{old}}$ compared to $\mathcal{C}_{\text{modern}}$. This pattern is primarily explained by the fact that these words possess multiple conventional, neutral usages across broad domains (agriculture, cooking, botany, weather, or everyday trade), resulting in widespread occurrences across two centuries of digitized newspaper journalism. Conversely, their weaponized derogatory senses are narrower and more straightforward, making them rarer in general discourse, coupled with the fact that several target terms are not extensively used in contemporary everyday vocabulary.
    \item \textbf{Modern Subcultural Amplification:} A small subset of terms displays an inverse trend, appearing more frequently in $\mathcal{C}_{\text{modern}}$ than in $\mathcal{C}_{\text{old}}$: \texttt{shill} (1,115 vs.\ 338), \texttt{faggot} (186 vs.\ 87), and \texttt{thuggish} (119 vs.\ 8). These exceptions reflect vocabulary whose prevalence escalated dramatically in late 20th- and 21st-century online communities, political message boards, and digital vernacular.
    \item \textbf{Unfiltered Surface Matching Caveat:} We emphasize that these numbers reflect raw, unlemmatized surface-token occurrences across corpus documents without sense disambiguation. Consequently, counts are susceptible to noise, including historical OCR misrecognitions, homographs, proper noun co-occurrences (e.g., \texttt{karen} as a given name), and capitalization variants. Nonetheless, they establish robust corpus grounding and verify that these terms were in active circulation across both historical and modern periods.
\end{enumerate}

%% file: custom.bib
@string{acl = {Association for Computational Linguistics}}

@string{anth = {https://aclanthology.org/}}

@article{deMauro16,
author = {De Mauro, Tullio},
title = {Le parole per ferire},
journal = {Internazionale},
volume={},
note = {Compiled for the ``Joe Cox" Committee on intolerance, xenophobia, racism and hate phenomena, of the Italian Chamber of Deputies},
year = {2016},
url= {https://www.internazionale.it/opinione/tullio-de-mauro/2016/09/27/razzismo-parole-ferire}
}

@book{cepollaro20,
    author = {Cepollaro, Bianca},
    title = {Slurs and Thick Terms: When Language Encodes Values},
    publisher = {Philosophy of Language: Connections and Perspectives. Philosophy of Language: Connections and Perspectives. Bloomsbury Collections},
    year = {2020},
    url = {http://dx.doi.org/10.5040/9781978728189}
}

@article{BorkowskaKleparski2007,
  author    = {Paulina Borkowska and Grzegorz Kleparski},
  title     = {It befalls words to fall down : pejoration as a type of semantic change},
  journal   = {Uniwersytet Rzeszowski: Zeszyty naukowe Uniwersytetu Rzeszowskiego / Seria filologiczna / Studia anglica Resoviensia},
  volume    = {4},
  number    = {47/2007},
  pages     = {33 -- 50},
  url       = {http://www.univ.rzeszow.pl/file/1294/sar_v4_03.pdf},
  year      = {2007},
}

@article{cepollaro2023successes,
  title={The successes of reclamation},
  author={Cepollaro, Bianca and L{\'o}pez de Sa, Dan},
  journal={Synthese},
  volume={202},
  number={6},
  pages={178},
  year={2023},
  publisher={Springer},
  doi={10.1007/s11229-023-04393-y}
}

@inproceedings{ferrando2026multipride,
  title={MultiPRIDE at EVALITA 2026: Overview of the Multilingual Automatic Detection of Slur Reclamation in the LGBTQ+ Context Task},
  author={Ferrando, Chiara and Draetta, Lia and Madeddu, Marco and Sosto, Mae and Patti, Viviana and Rosso, Paolo and Bosco, Cristina and Mata, Jacinto and Gualda, Estrella},
  booktitle={Proceedings of the 9th Evaluation Campaign of Natural Language Processing and Speech Tools for Italian. Final Workshop (EVALITA 2026)},
  series={CEUR Workshop Proceedings},
  volume={4195},
  publisher={CEUR-WS.org},
  year={2026},
  url={https://aclanthology.org/2026.evalita-1.18/}
}

@inproceedings{rezvan-etal-2018-quality, author = {Rezvan, Mohammadreza and Shekarpour, Saeedeh and Balasuriya, Lakshika and Thirunarayan, Krishnaprasad and Shalin, Valerie L. and Sheth, Amit},
title = {A Quality Type-aware Annotated Corpus and Lexicon for Harassment Research},
year = {2018},
isbn = {9781450355636},
publisher = {Association for Computing Machinery},
address = {New York, NY, USA},
url = {https://doi.org/10.1145/3201064.3201103},
doi = {10.1145/3201064.3201103},
booktitle = {Proceedings of the 10th ACM Conference on Web Science},
pages = {33–36},
numpages = {4},
location = {Amsterdam, Netherlands},
series = {WebSci '18}
}

@misc{dell-etal-2023-american,
      title={American Stories: A Large-Scale Structured Text Dataset of Historical U.S. Newspapers}, 
      author={Melissa Dell and Jacob Carlson and Tom Bryan and Emily Silcock and Abhishek Arora and Zejiang Shen and Luca D'Amico-Wong and Quan Le and Pablo Querubin and Leander Heldring},
      year={2023},
      eprint={2308.12477},
      archivePrefix={arXiv},
      primaryClass={cs.CL},
      url={https://arxiv.org/abs/2308.12477}, 
}

@misc{chakraborty-2023-normal,
	title={Normal Hate and Offensive Speeches},
	publisher={Kaggle},
	author={Subhadeep Chakraborty},
	year={2023},
    url = {https://www.kaggle.com/datasets/subhajournal/normal-hate-and-offensive-speeches}
}

@misc{jikeli-etal-2023-hate,
  author       = {Jikeli, Gunther and
                  Karali, Sameer and
                  Soemer, Katharina},
  title        = {Hate Speech and Bias against Asians, Blacks, Jews,
                   Latines, and Muslims: A Dataset for Machine
                   Learning and Text Analytics
                  },
  month        = mar,
  year         = 2023,
  publisher    = {Zenodo},
  doi          = {10.5281/zenodo.10812805},
  url          = {https://doi.org/10.5281/zenodo.10812805},
}

@article{kennedy-etal-2020-gab,
    author       = {Brendan Kennedy and
                  Mohammad Atari and
                  Aida Mostafazadeh Davani and
                  Leigh Yeh and
                  Ali Omrani and
                  Yehsong Kim and
                  Kris Coombs and
                  Shreya Havaldar and
                  Gwenyth Portillo{-}Wightman and
                  Elaine Gonzalez and
                  Joe Hoover and
                  Aida Azatian and
                  Alyzeh Hussain and
                  Austin Lara and
                  Gabriel Cardenas and
                  Adam Omary and
                  Christina Park and
                  Xin Wang and
                  Clarisa Wijaya and
                  Yong Zhang and
                  Beth Meyerowitz and
                  Morteza Dehghani},
  title        = {Introducing the Gab Hate Corpus: defining and applying hate-based
                  rhetoric to social media posts at scale},
  journal      = {Lang. Resour. Evaluation},
  volume       = {56},
  number       = {1},
  pages        = {79--108},
  year         = {2022},
  url          = {https://doi.org/10.1007/s10579-021-09569-x}
}

@misc{kaggle-user-2019-hatred,
  title        = {Hatred during Metoo Movement},
  author       = {Kaggle User},
  year         = 2019,
  url         = {https://www.kaggle.com/datasets/rahulgoel1106/hatred-on-twitter-during-metoo-movement}
}

@misc{vu-etal-2023-extremebb,
      title={ExtremeBB: A Database for Large-Scale Research into Online Hate, Harassment, the Manosphere and Extremism}, 
      author={Anh V. Vu and Lydia Wilson and Yi Ting Chua and Ilia Shumailov and Ross Anderson},
      year={2023},
      eprint={2111.04479},
      archivePrefix={arXiv},
      primaryClass={cs.SI},
      url={https://arxiv.org/abs/2111.04479}, 
}

@misc{yang-etal-2025-qwen3,
      title={Qwen3 Technical Report}, 
      author={An Yang and Anfeng Li and Baosong Yang and Beichen Zhang and Binyuan Hui and Bo Zheng and Bowen Yu and Chang Gao and Chengen Huang and Chenxu Lv and Chujie Zheng and Dayiheng Liu and Fan Zhou and Fei Huang and Feng Hu and Hao Ge and Haoran Wei and Huan Lin and Jialong Tang and Jian Yang and Jianhong Tu and Jianwei Zhang and Jianxin Yang and Jiaxi Yang and Jing Zhou and Jingren Zhou and Junyang Lin and Kai Dang and Keqin Bao and Kexin Yang and Le Yu and Lianghao Deng and Mei Li and Mingfeng Xue and Mingze Li and Pei Zhang and Peng Wang and Qin Zhu and Rui Men and Ruize Gao and Shixuan Liu and Shuang Luo and Tianhao Li and Tianyi Tang and Wenbiao Yin and Xingzhang Ren and Xinyu Wang and Xinyu Zhang and Xuancheng Ren and Yang Fan and Yang Su and Yichang Zhang and Yinger Zhang and Yu Wan and Yuqiong Liu and Zekun Wang and Zeyu Cui and Zhenru Zhang and Zhipeng Zhou and Zihan Qiu},
      year={2025},
      eprint={2505.09388},
      archivePrefix={arXiv},
      primaryClass={cs.CL},
      url={https://arxiv.org/abs/2505.09388}, 
}

@misc{gabriel-2018-google,
  author       = {Gabriel, Robert James},
  title        = {{Google Profanity Words List}},
  year         = {2018},
  howpublished = {\url{https://github.com/coffee-and-fun/google-profanity-words}},
  note         = {GitHub repository; accessed 2025-11-05}
}

@misc{hatebase-2022-hatebase,
  author       = {{Hatebase.org}},
  title        = {{Hatebase: The World's Largest Structured Repository of Regionalized Hate Speech}},
  year         = {2022},
  howpublished = {\url{https://hatebase.org}},
  note         = {Accessed 2025-10-22}
}

@misc{weaponized-word-2023-weaponized,
  author       = {{Weaponized Word}},
  title        = {{Weaponized Words Lexicon}},
  year         = {2023},
  howpublished = {\url{https://weaponizedword.org}},
  note         = {Online lexicon; accessed 2024-10-22}
}

@misc{wikimedia-foundation-2023-wiktionary,
  author       = {{Wikimedia Foundation}},
  title        = {{Wiktionary: Category:English derogatory terms}},
  year         = {2023},
  howpublished = {\url{https://en.wiktionary.org/wiki/Category:English_derogatory_terms}},
  note         = {Accessed 2025-11-03}
}

@misc{wordnik-2024-wordnik,
  author       = {{Wordnik}},
  title        = {{Wordnik Offensive Terms List}},
  year         = {2024},
  howpublished = {\url{https://www.wordnik.com/lists/offensive}},
  note         = {Accessed 2025-10-10}
}

@misc{shanahan-etal-2023-role,
      title={Role-Play with Large Language Models}, 
      author={Murray Shanahan and Kyle McDonell and Laria Reynolds},
      year={2023},
      eprint={2305.16367},
      archivePrefix={arXiv},
      primaryClass={cs.CL},
      url={https://arxiv.org/abs/2305.16367}, 
}

@misc{wu-etal-2024-thinking,
      title={Thinking LLMs: General Instruction Following with Thought Generation}, 
      author={Tianhao Wu and Janice Lan and Weizhe Yuan and Jiantao Jiao and Jason Weston and Sainbayar Sukhbaatar},
      year={2024},
      eprint={2410.10630},
      archivePrefix={arXiv},
      primaryClass={cs.CL},
      url={https://arxiv.org/abs/2410.10630}, 
}

@misc{grattafiori-etal-2024-llama,
      title={The Llama 3 Herd of Models}, 
      author={Aaron Grattafiori and Abhimanyu Dubey and Abhinav Jauhri and Abhinav Pandey and Abhishek Kadian and Ahmad Al-Dahle and Aiesha Letman and Akhil Mathur and Alan Schelten and Alex Vaughan and Amy Yang and Angela Fan and Anirudh Goyal and Anthony Hartshorn and Aobo Yang and Archi Mitra and Archie Sravankumar and Artem Korenev and Arthur Hinsvark and Arun Rao and Aston Zhang and Aurelien Rodriguez and Austen Gregerson and Ava Spataru and Baptiste Roziere and Bethany Biron and Binh Tang and Bobbie Chern and Charlotte Caucheteux and Chaya Nayak and Chloe Bi and Chris Marra and Chris McConnell and Christian Keller and Christophe Touret and Chunyang Wu and Corinne Wong and Cristian Canton Ferrer and Cyrus Nikolaidis and Damien Allonsius and Daniel Song and Danielle Pintz and Danny Livshits and Danny Wyatt and David Esiobu and Dhruv Choudhary and Dhruv Mahajan and Diego Garcia-Olano and Diego Perino and Dieuwke Hupkes and Egor Lakomkin and Ehab AlBadawy and Elina Lobanova and Emily Dinan and Eric Michael Smith and Filip Radenovic and Francisco Guzm{\'a}n and Frank Zhang and Gabriel Synnaeve and Gabrielle Lee and Georgia Lewis Anderson and Govind Thattai and Graeme Nail and Gregoire Mialon and Guan Pang and Guillem Cucurell and Hailey Nguyen and Hannah Korevaar and Hu Xu and Hugo Touvron and Iliyan Zarov and Imanol Arrieta Ibarra and Isabel Kloumann and Ishan Misra and Ivan Evtimov and Jack Zhang and Jade Copet and Jaewon Lee and Jan Geffert and Jana Vranes and Jason Park and Jay Mahadeokar and Jeet Shah and Jelmer van der Linde and Jennifer Billock and Jenny Hong and Jenya Lee and Jeremy Fu and Jianfeng Chi and Jianyu Huang and Jiawen Liu and Jie Wang and Jiecao Yu and Joanna Bitton and Joe Spisak and Jongsoo Park and Joseph Rocca and Joshua Johnstun and Joshua Saxe and Junteng Jia and Kalyan Vasuden Alwala and Karthik Prasad and Kartikeya Upasani and Kate Plawiak and Ke Li and Kenneth Heafield and Kevin Stone and Khalid El-Arini and Krithika Iyer and Kshitiz Malik and Kuenley Chiu and Kunal Bhalla and Kushal Lakhotia and Lauren Rantala-Yeary and Laurens van der Maaten and Lawrence Chen and Liang Tan and Liz Jenkins and Louis Martin and Lovish Madaan and Lubo Malo and Lukas Blecher and Lukas Landzaat and Luke de Oliveira and Madeline Muzzi and Mahesh Pasupuleti and Mannat Singh and Manohar Paluri and Marcin Kardas and Maria Tsimpoukelli and Mathew Oldham and Mathieu Rita and Maya Pavlova and Melanie Kambadur and Mike Lewis and Min Si and Mitesh Kumar Singh and Mona Hassan and Naman Goyal and Narjes Torabi and Nikolay Bashlykov and Nikolay Bogoychev and Niladri Chatterji and Ning Zhang and Olivier Duchenne and Onur {\c{C}}elebi and Patrick Alrassy and Pengchuan Zhang and Pengwei Li and Petar Vasic and Peter Weng and Prajjwal Bhargava and Pratik Dubal and Praveen Krishnan and Punit Singh Koura and Puxin Xu and Qing He and Qingxiao Dong and Ragavan Srinivasan and Raj Ganapathy and Ramon Calderer and Ricardo Silveira Cabral and Robert Stojnic and Roberta Raileanu and Rohan Maheswari and Rohit Girdhar and Rohit Patel and Romain Sauvestre and Ronnie Polidoro and Roshan Sumbaly and Ross Taylor and Ruan Silva and Rui Hou and Rui Wang and Saghar Hosseini and Sahana Chennabasappa and Sanjay Singh and Sean Bell and Seohyun Sonia Kim and Sergey Edunov and Shaoliang Nie and Sharan Narang and Sharath Raparthy and Sheng Shen and Shengye Wan and Shruti Bhosale and Shun Zhang and Simon Vandenhende and Soumya Batra and Spencer Whitman and Sten Sootla and Stephane Collot and Suchin Gururangan and Sydney Borodinsky and Tamar Herman and Tara Fowler and Tarek Sheasha and Thomas Georgiou and Thomas Scialom and Tobias Speckbacher and Todor Mihaylov and Tong Xiao and Ujjwal Karn and Vedanuj Goswami and Vibhor Gupta and Vignesh Ramanathan and Viktor Kerkez and Vincent Gonguet and Virginie Do and Vish Vogeti and V{\'i}tor Albiero and Vladan Petrovic and Weiwei Chu and Wenhan Xiong and Wenyin Fu and Whitney Meers and Xavier Martinet and Xiaodong Wang and Xiaofang Wang and Xiaoqing Ellen Tan and Xide Xia and Xinfeng Xie and Xuchao Jia and Xuewei Wang and Yaelle Goldschlag and Yashesh Gaur and Yasmine Babaei and Yi Wen and Yiwen Song and Yuchen Zhang and Yue Li and Yuning Mao and Zacharie Delpierre Coudert and Zheng Yan and Zhengxing Chen and Zoe Papakipos and Aaditya Singh and Aayushi Srivastava and Abha Jain and Adam Kelsey and Adam Shajnfeld and Adithya Gangidi and Adolfo Victoria and Ahuva Goldstand and Ajay Menon and Ajay Sharma and Alex Boesenberg and Alexei Baevski and Allie Feinstein and Amanda Kallet and Amit Sangani and Amos Teo and Anam Yunus and Andrei Lupu and Andres Alvarado and Andrew Caples and Andrew Gu and Andrew Ho and Andrew Poulton and Andrew Ryan and Ankit Ramchandani and Annie Dong and Annie Franco and Anuj Goyal and Aparajita Saraf and Arkabandhu Chowdhury and Ashley Gabriel and Ashwin Bharambe and Assaf Eisenman and Azadeh Yazdan and Beau James and Ben Maurer and Benjamin Leonhardi and Bernie Huang and Beth Loyd and Beto De Paola and Bhargavi Paranjape and Bing Liu and Bo Wu and Boyu Ni and Braden Hancock and Bram Wasti and Brandon Spence and Brani Stojkovic and Brian Gamido and Britt Montalvo and Carl Parker and Carly Burton and Catalina Mejia and Ce Liu and Changhan Wang and Changkyu Kim and Chao Zhou and Chester Hu and Ching-Hsiang Chu and Chris Cai and Chris Tindal and Christoph Feichtenhofer and Cynthia Gao and Damon Civin and Dana Beaty and Daniel Kreymer and Daniel Li and David Adkins and David Xu and Davide Testuggine and Delia David and Devi Parikh and Diana Liskovich and Didem Foss and Dingkang Wang and Duc Le and Dustin Holland and Edward Dowling and Eissa Jamil and Elaine Montgomery and Eleonora Presani and Emily Hahn and Emily Wood and Eric-Tuan Le and Erik Brinkman and Esteban Arcaute and Evan Dunbar and Evan Smothers and Fei Sun and Felix Kreuk and Feng Tian and Filippos Kokkinos and Firat Ozgenel and Francesco Caggioni and Frank Kanayet and Frank Seide and Gabriela Medina Florez and Gabriella Schwarz and Gada Badeer and Georgia Swee and Gil Halpern and Grant Herman and Grigory Sizov and Guangyi and Zhang and Guna Lakshminarayanan and Hakan Inan and Hamid Shojanazeri and Han Zou and Hannah Wang and Hanwen Zha and Haroun Habeeb and Harrison Rudolph and Helen Suk and Henry Aspegren and Hunter Goldman and Hongyuan Zhan and Ibrahim Damlaj and Igor Molybog and Igor Tufanov and Ilias Leontiadis and Irina-Elena Veliche and Itai Gat and Jake Weissman and James Geboski and James Kohli and Janice Lam and Japhet Asher and Jean-Baptiste Gaya and Jeff Marcus and Jeff Tang and Jennifer Chan and Jenny Zhen and Jeremy Reizenstein and Jeremy Teboul and Jessica Zhong and Jian Jin and Jingyi Yang and Joe Cummings and Jon Carvill and Jon Shepard and Jonathan McPhie and Jonathan Torres and Josh Ginsburg and Junjie Wang and Kai Wu and Kam Hou U and Karan Saxena and Kartikay Khandelwal and Katayoun Zand and Kathy Matosich and Kaushik Veeraraghavan and Kelly Michelena and Keqian Li and Kiran Jagadeesh and Kun Huang and Kunal Chawla and Kyle Huang and Lailin Chen and Lakshya Garg and Lavender A and Leandro Silva and Lee Bell and Lei Zhang and Liangpeng Guo and Licheng Yu and Liron Moshkovich and Luca Wehrstedt and Madian Khabsa and Manav Avalani and Manish Bhatt and Martynas Mankus and Matan Hasson and Matthew Lennie and Matthias Reso and Maxim Groshev and Maxim Naumov and Maya Lathi and Meghan Keneally and Miao Liu and Michael L. Seltzer and Michal Valko and Michelle Restrepo and Mihir Patel and Mik Vyatskov and Mikayel Samvelyan and Mike Clark and Mike Macey and Mike Wang and Miquel Jubert Hermoso and Mo Metanat and Mohammad Rastegari and Munish Bansal and Nandhini Santhanam and Natascha Parks and Natasha White and Navyata Bawa and Nayan Singhal and Nick Egebo and Nicolas Usunier and Nikhil Mehta and Nikolay Pavlovich Laptev and Ning Dong and Norman Cheng and Oleg Chernoguz and Olivia Hart and Omkar Salpekar and Ozlem Kalinli and Parkin Kent and Parth Parekh and Paul Saab and Pavan Balaji and Pedro Rittner and Philip Bontrager and Pierre Roux and Piotr Dollar and Polina Zvyagina and Prashant Ratanchandani and Pritish Yuvraj and Qian Liang and Rachad Alao and Rachel Rodriguez and Rafi Ayub and Raghotham Murthy and Raghu Nayani and Rahul Mitra and Rangaprabhu Parthasarathy and Raymond Li and Rebekkah Hogan and Robin Battey and Rocky Wang and Russ Howes and Ruty Rinott and Sachin Mehta and Sachin Siby and Sai Jayesh Bondu and Samyak Datta and Sara Chugh and Sara Hunt and Sargun Dhillon and Sasha Sidorov and Satadru Pan and Saurabh Mahajan and Saurabh Verma and Seiji Yamamoto and Sharadh Ramaswamy and Shaun Lindsay and Shaun Lindsay and Sheng Feng and Shenghao Lin and Shengxin Cindy Zha and Shishir Patil and Shiva Shankar and Shuqiang Zhang and Shuqiang Zhang and Sinong Wang and Sneha Agarwal and Soji Sajuyigbe and Soumith Chintala and Stephanie Max and Stephen Chen and Steve Kehoe and Steve Satterfield and Sudarshan Govindaprasad and Sumit Gupta and Summer Deng and Sungmin Cho and Sunny Virk and Suraj Subramanian and Sy Choudhury and Sydney Goldman and Tal Remez and Tamar Glaser and Tamara Best and Thilo Koehler and Thomas Robinson and Tianhe Li and Tianjun Zhang and Tim Matthews and Timothy Chou and Tzook Shaked and Varun Vontimitta and Victoria Ajayi and Victoria Montanez and Vijai Mohan and Vinay Satish Kumar and Vishal Mangla and Vlad Ionescu and Vlad Poenaru and Vlad Tiberiu Mihailescu and Vladimir Ivanov and Wei Li and Wenchen Wang and Wenwen Jiang and Wes Bouaziz and Will Constable and Xiaocheng Tang and Xiaojian Wu and Xiaolan Wang and Xilun Wu and Xinbo Gao and Yaniv Kleinman and Yanjun Chen and Ye Hu and Ye Jia and Ye Qi and Yenda Li and Yilin Zhang and Ying Zhang and Yossi Adi and Youngjin Nam and Yu and Wang and Yu Zhao and Yuchen Hao and Yundi Qian and Yunlu Li and Yuzi He and Zach Rait and Zachary DeVito and Zef Rosnbrick and Zhaoduo Wen and Zhenyu Yang and Zhiwei Zhao and Zhiyu Ma},
      year={2024},
      eprint={2407.21783},
      archivePrefix={arXiv},
      primaryClass={cs.AI},
      url={https://arxiv.org/abs/2407.21783}, 
}

@misc{kolli-etal-2026-word,
      title={Word-Centered Semantic Graphs for Interpretable Diachronic Sense Tracking}, 
      author={Imene Kolli and Kai-Robin Lange and Jonas Rieger and Carsten Jentsch},
      year={2026},
      eprint={2601.22410},
      archivePrefix={arXiv},
      primaryClass={cs.CL},
      url={https://arxiv.org/abs/2601.22410}, 
}

@Article{florio-etal-2020-time,
    AUTHOR = {Florio, Komal and Basile, Valerio and Polignano, Marco and Basile, Pierpaolo and Patti, Viviana},
    TITLE = {Time of Your Hate: The Challenge of Time in Hate Speech Detection on Social Media},
    JOURNAL = {Applied Sciences},
    VOLUME = {10},
    YEAR = {2020},
    NUMBER = {12},
    ARTICLE-NUMBER = {4180},
    URL = {https://www.mdpi.com/2076-3417/10/12/4180},
    ISSN = {2076-3417},
    DOI = {10.3390/app10124180}
}

@article{frenda-etal-2023-killing,
author = {Frenda, Simona and Patti, Viviana and Rosso, Paolo},
year = {2022},
month = {08},
pages = {1516-1537},
title = {Killing me softly: Creative and cognitive aspects of implicitness in abusive language online},
volume = {29},
journal = {Natural Language Engineering},
doi = {10.1017/S1351324922000316}
}

@article{poletto-etal-2021-resources,
  title={Resources and benchmark corpora for hate speech detection: a systematic review},
  author={Fabio Poletto and Valerio Basile and Manuela Sanguinetti and Cristina Bosco and Viviana Patti},
  journal={Language Resources and Evaluation},
  year={2020},
  volume={55},
  pages={477 - 523},
  url={https://api.semanticscholar.org/CorpusID:224846337}
}

@inproceedings{nobata-etal-2016-abusive,
author = {Nobata, Chikashi and Tetreault, Joel and Thomas, Achint and Mehdad, Yashar and Chang, Yi},
title = {Abusive Language Detection in Online User Content},
year = {2016},
isbn = {9781450341431},
publisher = {International World Wide Web Conferences Steering Committee},
address = {Republic and Canton of Geneva, CHE},
url = {https://doi.org/10.1145/2872427.2883062},
doi = {10.1145/2872427.2883062},
booktitle = {Proceedings of the 25th International Conference on World Wide Web},
pages = {145–153},
numpages = {9},
location = {Montr{\'e}al, Qu{\'e}bec, Canada},
series = {WWW '16}
}

@article{periti-montanelli-2024-lexical,
author = {Periti, Francesco and Montanelli, Stefano},
title = {Lexical Semantic Change through Large Language Models: a Survey},
year = {2024},
issue_date = {November 2024},
publisher = {Association for Computing Machinery},
address = {New York, NY, USA},
volume = {56},
number = {11},
issn = {0360-0300},
url = {https://doi.org/10.1145/3672393},
doi = {10.1145/3672393},
journal = {ACM Comput. Surv.},
month = jun,
articleno = {282},
numpages = {38}
}

@inproceedings{founta-etal-2018-large,
      title={Large Scale Crowdsourcing and Characterization of Twitter Abusive Behavior}, 
      author={Founta, Antigoni-Maria and Djouvas, Constantinos and Chatzakou, Despoina and Leontiadis, Ilias and Blackburn, Jeremy and Stringhini, Gianluca and Vakali, Athena and Sirivianos, Michael and Kourtellis, Nicolas},
    booktitle={11th International Conference on Web and Social Media, ICWSM 2018},
    year={2018},
    organization={AAAI Press},
    url = {https://doi.org/10.1609/icwsm.v12i1.14991} 
}

@inproceedings{borkan-etal-2019-nuanced,
author =     {Borkan, Daniel and Dixon, Lucas and Sorensen, Jeffrey and Thain, Nithum and Vasserman, Lucy},
title = {Nuanced Metrics for Measuring Unintended Bias with Real Data for Text Classification},
year = {2019},
isbn = {9781450366755},
publisher = {Association for Computing Machinery},
address = {New York, NY, USA},
url = {https://doi.org/10.1145/3308560.3317593},
doi = {10.1145/3308560.3317593},
booktitle = {Companion Proceedings of The 2019 World Wide Web Conference},
pages = {491–500},
numpages = {10},
location = {San Francisco, USA},
series = {WWW '19}
}

@inproceedings{davidson-etal-2017-automated,
      title={Automated Hate Speech Detection and the Problem of Offensive Language}, 
      author = {Davidson, Thomas and Warmsley, Dana and Macy, Michael and Weber, Ingmar}, 
  booktitle = {Proceedings of the 11th International AAAI Conference on Web and Social Media},
  series = {ICWSM '17},
  year = {2017},
  location = {Montreal, Canada},
  pages = {512-515}
  }

@misc{ali-etal-2025-evolving,
      title={Evolving Hate Speech Online: An Adaptive Framework for Detection and Mitigation}, 
      author={Shiza Ali and Jeremy Blackburn and Gianluca Stringhini},
      year={2025},
      eprint={2502.10921},
      archivePrefix={arXiv},
      primaryClass={cs.CL},
      url={https://arxiv.org/abs/2502.10921}, 
}

@misc{hegde-etal-2025-chronoberg,
      title={CHRONOBERG: Capturing Language Evolution and Temporal Awareness in Foundation Models}, 
      author={Niharika Hegde and Subarnaduti Paul and Lars Joel-Frey and Manuel Brack and Kristian Kersting and Martin Mundt and Patrick Schramowski},
      year={2025},
      eprint={2509.22360},
      archivePrefix={arXiv},
      primaryClass={cs.CL},
      url={https://arxiv.org/abs/2509.22360}, 
}

@misc{tahmasebi-etal-2027-semeval,
      title={SemEval-2027 Task 3: Semantic Change Across Multiple Time Periods}, 
      author={Nina Tahmasebi and Pierluigi Cassotti and Felix Morger and Lucia Siciliani and Eduardo Cal{\`o} and Pablo Mosteiro and Stefano De Pascale and Mariia Fedorova},
      year={2027},
      howpublished={\url{https://www.changeiskey.org/event/2027-semeval/}}, 
}

@misc{ciletti-2026-inspicio,
      title={Inspicio: Open-Vocabulary, LLM-Based Sense Retrieval for Historical Languages}, 
      author={Michele Ciletti},
      year={2026},
      eprint={2609.00998},
      archivePrefix={arXiv},
      primaryClass={cs.CL},
      url={https://arxiv.org/abs/2609.00998}, 
}

@misc{phan-tat-etal-2026-synflow,
      title={SynFlow: A Multidimensional Diachronic Semantic Analysis Toolkit}, 
      author={Bach Phan-Tat and Kris Heylen and Dirk Geeraerts and Stefano De Pascale and Dirk Speelman},
      year={2026},
      eprint={2608.19472},
      archivePrefix={arXiv},
      primaryClass={cs.CL},
      url={https://arxiv.org/abs/2608.19472}, 
}

@misc{yitzchak-etal-2026-words,
      title={When Words Divide: Diachronic Ideological Polarization in Political Discourse on Social Media}, 
      author={Roy Yitzchak and Noa Lavie and Ella Rabinovich},
      year={2026},
      eprint={2608.01176},
      archivePrefix={arXiv},
      primaryClass={cs.CL},
      url={https://arxiv.org/abs/2608.01176}, 
}

@misc{qureshi-etal-2026-training,
      title={Training-Time Explainability for Multilingual Hate Speech Detection: Aligning Model Reasoning with Human Rationales}, 
      author={Muhammad Deedahwar Mazhar Qureshi and Sannaan Khan and Muhammad Atif Qureshi and Wael Rashwan},
      year={2026},
      eprint={2608.26125},
      archivePrefix={arXiv},
      primaryClass={cs.CL},
      url={https://arxiv.org/abs/2608.26125}, 
}

@misc{fittschen-etal-2025-pretraining,
title={Pretraining Language Models for Diachronic Linguistic Change Discovery},
   url={http://dx.doi.org/10.18653/v1/2026.findings-eacl.241},
   DOI={10.18653/v1/2026.findings-eacl.241},
   booktitle={Findings of the Association for Computational Linguistics: EACL 2026},
   publisher={Association for Computational Linguistics},
   author={Fittschen, Elisabeth and Li, Sabrina Xin and Lippincott, Tom and Choshen, Leshem and Messner, Craig},
   year={2026},
   pages={4627–4642} }

@misc{jayatilleke-de-silva-2026-dynamics,
      title={Dynamics of Meaning: Towards the Evaluation of Diachronic Semantic Change in Sinhala}, 
      author={Nevidu Jayatilleke and Nisansa de Silva},
      year={2026},
      eprint={2609.08609},
      archivePrefix={arXiv},
      primaryClass={cs.CL},
      url={https://arxiv.org/abs/2609.08609}, 
}

@article{mollas-etal-2022-ethos,
    title={ETHOS: a multi-label hate speech detection dataset},
   volume={8},
   ISSN={2198-6053},
   url={http://dx.doi.org/10.1007/s40747-021-00608-2},
   DOI={10.1007/s40747-021-00608-2},
   number={6},
   journal={Complex \&; Intelligent Systems},
   publisher={Springer Science and Business Media LLC},
   author={Mollas, Ioannis and Chrysopoulou, Zoe and Karlos, Stamatis and Tsoumakas, Grigorios},
   year={2022},
   month=Jan, pages={4663–4678} }

@inproceedings{basile-etal-2019-semeval,title = "{S}em{E}val-2019 Task 5: Multilingual Detection of Hate Speech Against Immigrants and Women in {T}witter",author = "Basile, Valerio and Bosco, Cristina and Fersini, Elisabetta and Nozza, Debora and Patti, Viviana and Rangel Pardo, Francisco Manuel and Rosso, Paolo and Sanguinetti, Manuela",editor = "May, Jonathan and Shutova, Ekaterina and Herbelot, Aurelie and Zhu, Xiaodan and Apidianaki, Marianna and Mohammad, Saif M.",booktitle = "Proceedings of the 13th International Workshop on Semantic Evaluation",month = jun,year = "2019",address = "Minneapolis, Minnesota, USA",publisher = acl,url = anth # {S19-2007/},doi = "10.18653/v1/S19-2007",pages = "54--63"}

@inproceedings{bassignana-etal-2018-hurtlex,title = "Hurtlex: A Multilingual Lexicon of Words to Hurt",author = "Bassignana, Elisa and Basile, Valerio and Patti, Viviana",editor = "Cabrio, Elena and Mazzei, Alessandro and Tamburini, Fabio",booktitle = "Proceedings of the Fifth Italian Conference on Computational Linguistics (CLiC-it 2018)",month = dec,year = "2018",address = "Turin, Italy",publisher = "CEUR Workshop Proceedings",url = anth # {2018.clicit-1.11/},pages = "52--57",ISBN = "978-88-31978-41-5"}

@inproceedings{bond-choo-2021-taboo,title = "Taboo {W}ordnet",author = "Bond, Francis and Choo, Merrick Yeu Herng",editor = "Vossen, Piek and Fellbaum, Christiane",booktitle = "Proceedings of the 11th Global Wordnet Conference",month = jan,year = "2021",address = "University of South Africa (UNISA)",publisher = "Global Wordnet Association",url = anth # {2021.gwc-1.5/},doi = "10.18653/v1/2021.gwc-1.5",pages = "36--43"}

@inproceedings{bourgeade-etal-2023-multilingual,title = "A Multilingual Dataset of Racial Stereotypes in Social Media Conversational Threads",author = "Bourgeade, Tom and Cignarella, Alessandra Teresa and Frenda, Simona and Laurent, Mario and Schmeisser-Nieto, Wolfgang and Benamara, Farah and Bosco, Cristina and Moriceau, V{\'e}ronique and Patti, Viviana and Taul{\'e}, Mariona",editor = "Vlachos, Andreas and Augenstein, Isabelle",booktitle = "Findings of the Association for Computational Linguistics: EACL 2023",month = may,year = "2023",address = "Dubrovnik, Croatia",publisher = acl,url = anth # {2023.findings-eacl.51/},doi = "10.18653/v1/2023.findings-eacl.51",pages = "686--696"}

@inproceedings{d-zamora-reina-etal-2022-black,title = "{LSCD}iscovery: A shared task on semantic change discovery and detection in {S}panish",author = "Zamora-Reina, Frank D. and Bravo-Marquez, Felipe and Schlechtweg, Dominik",editor = "Tahmasebi, Nina and Montariol, Syrielle and Kutuzov, Andrey and Hengchen, Simon and Dubossarsky, Haim and Borin, Lars",booktitle = "Proceedings of the 3rd Workshop on Computational Approaches to Historical Language Change",month = may,year = "2022",address = "Dublin, Ireland",publisher = acl,url = anth # {2022.lchange-1.16/},doi = "10.18653/v1/2022.lchange-1.16",pages = "149--164"}

@inproceedings{dubossarsky-etal-2019-time,title = "Time-Out: Temporal Referencing for Robust Modeling of Lexical Semantic Change",author = "Dubossarsky, Haim and Hengchen, Simon and Tahmasebi, Nina and Schlechtweg, Dominik",editor = "Korhonen, Anna and Traum, David and M{\`a}rquez, Llu{\'i}s",booktitle = "Proceedings of the 57th Annual Meeting of the Association for Computational Linguistics",month = jul,year = "2019",address = "Florence, Italy",publisher = acl,url = anth # {P19-1044/},doi = "10.18653/v1/P19-1044",pages = "457--470"}

@inproceedings{fedorova-etal-2024-definition,title = "Definition generation for lexical semantic change detection",author = "Fedorova, Mariia and Kutuzov, Andrey and Scherrer, Yves",editor = "Ku, Lun-Wei and Martins, Andre and Srikumar, Vivek",booktitle = "Findings of the Association for Computational Linguistics: ACL 2024",month = aug,year = "2024",address = "Bangkok, Thailand",publisher = acl,url = anth # {2024.findings-acl.339/},doi = "10.18653/v1/2024.findings-acl.339",pages = "5712--5724"}

@inproceedings{fedorova-etal-2025-explaining,title = "Explaining novel senses using definition generation with open language models",author = "Fedorova, Mariia and Kutuzov, Andrey and Periti, Francesco and Scherrer, Yves",editor = "Christodoulopoulos, Christos and Chakraborty, Tanmoy and Rose, Carolyn and Peng, Violet",booktitle = "Findings of the Association for Computational Linguistics: EMNLP 2025",month = nov,year = "2025",address = "Suzhou, China",publisher = acl,url = anth # {2025.findings-emnlp.1214/},doi = "10.18653/v1/2025.findings-emnlp.1214",pages = "22294--22302",ISBN = "979-8-89176-335-7"}

@inproceedings{fedorova-etal-2026-dhplt,title = "{DHPLT}: large-scale multilingual diachronic corpora and word representations for semantic change modelling",author = "Fedorova, Mariia and Kutuzov, Andrey and Umarova, Khonzoda",editor = "Tahmasebi, Nina and Cassotti, Pierluigi and Montariol, Syrielle and Kutuzov, Andrey and Huebscher, Netta and Spaziani, Elena and Baes, Naomi",booktitle = "The Proceedings for the 6th International Workshop on Computational Approaches to Language Change ({LC}hange{'}26)",month = mar,year = "2026",address = "Rabat, Morocco",publisher = acl,url = anth # {2026.lchange-1.7/},doi = "10.18653/v1/2026.lchange-1.7",pages = "87--96",ISBN = "979-8-89176-362-3"}

@inproceedings{giulianelli-etal-2020-analysing,title = "Analysing Lexical Semantic Change with Contextualised Word Representations",author = "Giulianelli, Mario and Del Tredici, Marco and Fern{\'a}ndez, Raquel",editor = "Jurafsky, Dan and Chai, Joyce and Schluter, Natalie and Tetreault, Joel",booktitle = "Proceedings of the 58th Annual Meeting of the Association for Computational Linguistics",month = jul,year = "2020",address = "Online",publisher = acl,url = anth # {2020.acl-main.365/},doi = "10.18653/v1/2020.acl-main.365",pages = "3960--3973"}

@inproceedings{hamilton-etal-2016-diachronic,title = "Diachronic Word Embeddings Reveal Statistical Laws of Semantic Change",author = "Hamilton, William L. and Leskovec, Jure and Jurafsky, Dan",editor = "Erk, Katrin and Smith, Noah A.",booktitle = "Proceedings of the 54th Annual Meeting of the Association for Computational Linguistics (Volume 1: Long Papers)",month = aug,year = "2016",address = "Berlin, Germany",publisher = acl,url = anth # {P16-1141/},doi = "10.18653/v1/P16-1141",pages = "1489--1501"}

@inproceedings{hoeken-etal-2023-towards-detecting,title = "Towards Detecting Lexical Change of Hate Speech in Historical Data",author = {Hoeken, Sanne and Spliethoff, Sophie and Schwandt, Silke and Zarrie{\ss}, Sina and Alacam, {\"O}zge},editor = "Tahmasebi, Nina and Montariol, Syrielle and Dubossarsky, Haim and Kutuzov, Andrey and Hengchen, Simon and Alfter, David and Periti, Francesco and Cassotti, Pierluigi",booktitle = "Proceedings of the 4th Workshop on Computational Approaches to Historical Language Change",month = dec,year = "2023",address = "Singapore",publisher = acl,url = anth # {2023.lchange-1.11/},doi = "10.18653/v1/2023.lchange-1.11",pages = "100--111"}

@inproceedings{kiyama-etal-2025-analyzing,title = "Analyzing Continuous Semantic Shifts with Diachronic Word Similarity Matrices",author = "Kiyama, Hajime and Aida, Taichi and Komachi, Mamoru and Ogiso, Toshinobu and Takamura, Hiroya and Mochihashi, Daichi",editor = "Rambow, Owen and Wanner, Leo and Apidianaki, Marianna and Al-Khalifa, Hend and Eugenio, Barbara Di and Schockaert, Steven",booktitle = "Proceedings of the 31st International Conference on Computational Linguistics",month = jan,year = "2025",address = "Abu Dhabi, UAE",publisher = acl,url = anth # {2025.coling-main.109/},pages = "1613--1631"}

@inproceedings{laicher-etal-2021-explaining,title = "Explaining and Improving {BERT} Performance on Lexical Semantic Change Detection",author = "Laicher, Severin and Kurtyigit, Sinan and Schlechtweg, Dominik and Kuhn, Jonas and Schulte im Walde, Sabine",editor = "Sorodoc, Ionut-Teodor and Sushil, Madhumita and Takmaz, Ece and Agirre, Eneko",booktitle = "Proceedings of the 16th Conference of the European Chapter of the Association for Computational Linguistics: Student Research Workshop",month = apr,year = "2021",address = "Online",publisher = acl,url = anth # {2021.eacl-srw.25/},doi = "10.18653/v1/2021.eacl-srw.25",pages = "192--202"}

@inproceedings{liu-etal-2026-beyond,title = "Beyond Supervised Clarification: Input Rewriting with {LLM}s for Dialogue Discourse Parsing",author = "Liu, Yiming and Zhang, Ziyue and Xu, Zhichao and Yu, Xin and Tang, Yingheng and Jiang, Tianyu and Cao, Jie",editor = "Choi, Jinho D. and Chen, Yun-Nung and Funakoshi, Kotaro and Emami, Ali",booktitle = "Proceedings of the 27th Annual Meeting of the Special Interest Group on Discourse and Dialogue",month = aug,year = "2026",address = "Atlanta, Georgia, USA",publisher = acl,url = anth # {2026.sigdial-1.30/},doi = "10.18653/v1/2026.sigdial-1.30",pages = "423--439"}

@inproceedings{loureiro-etal-2022-tempowic,title = "{T}empo{W}i{C}: An Evaluation Benchmark for Detecting Meaning Shift in Social Media",author = "Loureiro, Daniel and D{'}Souza, Aminette and Muhajab, Areej Nasser and White, Isabella A. and Wong, Gabriel and Espinosa-Anke, Luis and Neves, Leonardo and Barbieri, Francesco and Camacho-Collados, Jose",editor = "Calzolari, Nicoletta and Huang, Chu-Ren and Kim, Hansaem and Pustejovsky, James and Wanner, Leo and Choi, Key-Sun and Ryu, Pum-Mo and Chen, Hsin-Hsi and Donatelli, Lucia and Ji, Heng and Kurohashi, Sadao and Paggio, Patrizia and Xue, Nianwen and Kim, Seokhwan and Hahm, Younggyun and He, Zhong and Lee, Tony Kyungil and Santus, Enrico and Bond, Francis and Na, Seung-Hoon",booktitle = "Proceedings of the 29th International Conference on Computational Linguistics",month = oct,year = "2022",address = "Gyeongju, Republic of Korea",publisher = "International Committee on Computational Linguistics",url = anth # {2022.coling-1.296/},pages = "3353--3359"}

@inproceedings{magu-luo-2018-determining,title = "Determining Code Words in Euphemistic Hate Speech Using Word Embedding Networks",author = "Magu, Rijul and Luo, Jiebo",editor = "Fi{\v{s}}er, Darja and Huang, Ruihong and Prabhakaran, Vinodkumar and Voigt, Rob and Waseem, Zeerak and Wernimont, Jacqueline",booktitle = "Proceedings of the 2nd Workshop on Abusive Language Online ({ALW}2)",month = oct,year = "2018",address = "Brussels, Belgium",publisher = acl,url = anth # {W18-5112/},doi = "10.18653/v1/W18-5112",pages = "93--100"}

@inproceedings{montariol-etal-2021-scalable,title = "Scalable and Interpretable Semantic Change Detection",author = "Montariol, Syrielle and Martinc, Matej and Pivovarova, Lidia",editor = "Toutanova, Kristina and Rumshisky, Anna and Zettlemoyer, Luke and Hakkani-Tur, Dilek and Beltagy, Iz and Bethard, Steven and Cotterell, Ryan and Chakraborty, Tanmoy and Zhou, Yichao",booktitle = "Proceedings of the 2021 Conference of the North American Chapter of the Association for Computational Linguistics: Human Language Technologies",month = jun,year = "2021",address = "Online",publisher = acl,url = anth # {2021.naacl-main.369/},doi = "10.18653/v1/2021.naacl-main.369",pages = "4642--4652"}

@inproceedings{nozza-hovy-2023-state,title = "The State of Profanity Obfuscation in Natural Language Processing Scientific Publications",author = "Nozza, Debora and Hovy, Dirk",editor = "Rogers, Anna and Boyd-Graber, Jordan and Okazaki, Naoaki",booktitle = "Findings of the Association for Computational Linguistics: ACL 2023",month = jul,year = "2023",address = "Toronto, Canada",publisher = acl,url = anth # {2023.findings-acl.240/},doi = "10.18653/v1/2023.findings-acl.240",pages = "3897--3909"}

@inproceedings{sadvilkar-neumann-2020-pysbd,title = "{P}y{SBD}: Pragmatic Sentence Boundary Disambiguation",author = "Sadvilkar, Nipun and Neumann, Mark",editor = "Park, Eunjeong L. and Hagiwara, Masato and Milajevs, Dmitrijs and Liu, Nelson F. and Chauhan, Geeticka and Tan, Liling",booktitle = "Proceedings of Second Workshop for NLP Open Source Software (NLP-OSS)",month = nov,year = "2020",address = "Online",publisher = acl,url = anth # {2020.nlposs-1.15/},doi = "10.18653/v1/2020.nlposs-1.15",pages = "110--114"}

@inproceedings{schlechtweg-etal-2020-semeval,title = "{S}em{E}val-2020 Task 1: Unsupervised Lexical Semantic Change Detection",author = "Schlechtweg, Dominik and McGillivray, Barbara and Hengchen, Simon and Dubossarsky, Haim and Tahmasebi, Nina",editor = "Herbelot, Aurelie and Zhu, Xiaodan and Palmer, Alexis and Schneider, Nathan and May, Jonathan and Shutova, Ekaterina",booktitle = "Proceedings of the Fourteenth Workshop on Semantic Evaluation",month = dec,year = "2020",address = "Barcelona (online)",publisher = "International Committee for Computational Linguistics",url = anth # {2020.semeval-1.1/},doi = "10.18653/v1/2020.semeval-1.1",pages = "1--23"}

@inproceedings{schlechtweg-etal-2026-lscd,title = "The {LSCD} Benchmark: a Testbed for Diachronic Word Meaning Tasks",author = "Schlechtweg, Dominik and Yadav, Sachin and Kuhn, Jonas and Arefyev, Nikolay",editor = "Mohammad, Saif M. and Ousidhoum, Nedjma",booktitle = "Proceedings of the 15th Joint Conference on Lexical and Computational Semantics (*{SEM} 2026)",month = jul,year = "2026",address = "San Diego, California, United States",publisher = acl,url = anth # {2026.starsem-conference.10/},doi = "10.18653/v1/2026.starsem-conference.10",pages = "148--163",ISBN = "979-8-89176-413-2"}

@inproceedings{schmidt-wiegand-2017-survey,title = "A Survey on Hate Speech Detection using Natural Language Processing",author = "Schmidt, Anna and Wiegand, Michael",editor = "Ku, Lun-Wei and Li, Cheng-Te",booktitle = "Proceedings of the Fifth International Workshop on Natural Language Processing for Social Media",month = apr,year = "2017",address = "Valencia, Spain",publisher = acl,url = anth # {W17-1101/},doi = "10.18653/v1/W17-1101",pages = "1--10"}

@inproceedings{tat-etal-2026-reframe,title = "{R}e{FRAME} or Remain: Unsupervised Lexical Semantic Change Detection with Frame Semantics",author = "Tat, Bach Phan and Heylen, Kris and De Pascale, Stefano and Geeraerts, Dirk and Speelman, Dirk",editor = "Mohammad, Saif M. and Ousidhoum, Nedjma",booktitle = "Proceedings of the 15th Joint Conference on Lexical and Computational Semantics (*{SEM} 2026)",month = jul,year = "2026",address = "San Diego, California, United States",publisher = acl,url = anth # {2026.starsem-conference.5/},doi = "10.18653/v1/2026.starsem-conference.5",pages = "83--97",ISBN = "979-8-89176-413-2"}

@inproceedings{toraman-etal-2022-large,title = "Large-Scale Hate Speech Detection with Cross-Domain Transfer",author = "Toraman, Cagri and {\c{S}}ahinu{\c{c}}, Furkan and Yilmaz, Eyup",editor = "Calzolari, Nicoletta and B{\'e}chet, Fr{\'e}d{\'e}ric and Blache, Philippe and Choukri, Khalid and Cieri, Christopher and Declerck, Thierry and Goggi, Sara and Isahara, Hitoshi and Maegaard, Bente and Mariani, Joseph and Mazo, H{\'e}l{\`e}ne and Odijk, Jan and Piperidis, Stelios",booktitle = "Proceedings of the Thirteenth Language Resources and Evaluation Conference",month = jun,year = "2022",address = "Marseille, France",publisher = "European Language Resources Association",url = anth # {2022.lrec-1.238/},pages = "2215--2225"}

@inproceedings{tseng-etal-2024-two,title = "Two Tales of Persona in {LLM}s: A Survey of Role-Playing and Personalization",author = "Tseng, Yu-Min and Huang, Yu-Chao and Hsiao, Teng-Yun and Chen, Wei-Lin and Huang, Chao-Wei and Meng, Yu and Chen, Yun-Nung",editor = "Al-Onaizan, Yaser and Bansal, Mohit and Chen, Yun-Nung",booktitle = "Findings of the Association for Computational Linguistics: EMNLP 2024",month = nov,year = "2024",address = "Miami, Florida, USA",publisher = acl,url = anth # {2024.findings-emnlp.969/},doi = "10.18653/v1/2024.findings-emnlp.969",pages = "16612--16631"}

@inproceedings{vidgen-etal-2019-challenges,title = "Challenges and frontiers in abusive content detection",author = "Vidgen, Bertie and Harris, Alex and Nguyen, Dong and Tromble, Rebekah and Hale, Scott and Margetts, Helen",editor = "Roberts, Sarah T. and Tetreault, Joel and Prabhakaran, Vinodkumar and Waseem, Zeerak",booktitle = "Proceedings of the Third Workshop on Abusive Language Online",month = aug,year = "2019",address = "Florence, Italy",publisher = acl,url = anth # {W19-3509/},doi = "10.18653/v1/W19-3509",pages = "80--93"}

@inproceedings{waseem-hovy-2016-hateful,title = "Hateful Symbols or Hateful People? Predictive Features for Hate Speech Detection on {T}witter",author = "Waseem, Zeerak and Hovy, Dirk",editor = "Andreas, Jacob and Choi, Eunsol and Lazaridou, Angeliki",booktitle = "Proceedings of the {NAACL} Student Research Workshop",month = jun,year = "2016",address = "San Diego, California",publisher = acl,url = anth # {N16-2013/},doi = "10.18653/v1/N16-2013",pages = "88--93"}

@inproceedings{wiegand-etal-2018-inducing,title = "Inducing a Lexicon of Abusive Words {--} a Feature-Based Approach",author = "Wiegand, Michael and Ruppenhofer, Josef and Schmidt, Anna and Greenberg, Clayton",editor = "Walker, Marilyn and Ji, Heng and Stent, Amanda",booktitle = "Proceedings of the 2018 Conference of the North {A}merican Chapter of the Association for Computational Linguistics: Human Language Technologies, Volume 1 (Long Papers)",month = jun,year = "2018",address = "New Orleans, Louisiana",publisher = acl,url = anth # {N18-1095/},doi = "10.18653/v1/N18-1095",pages = "1046--1056"}

@inproceedings{wiegand-etal-2019-detection,title = "{D}etection of {A}busive {L}anguage: the {P}roblem of {B}iased {D}atasets",author = "Wiegand, Michael and Ruppenhofer, Josef and Kleinbauer, Thomas",editor = "Burstein, Jill and Doran, Christy and Solorio, Thamar",booktitle = "Proceedings of the 2019 Conference of the North {A}merican Chapter of the Association for Computational Linguistics: Human Language Technologies, Volume 1 (Long and Short Papers)",month = jun,year = "2019",address = "Minneapolis, Minnesota",publisher = acl,url = anth # {N19-1060/},doi = "10.18653/v1/N19-1060",pages = "602--608"}

@inproceedings{zhang-etal-2025-llm,title = "{LLM} Sensitivity Challenges in Abusive Language Detection: Instruction-Tuned vs. Human Feedback",author = "Zhang, Yaqi and Hangya, Viktor and Fraser, Alexander",editor = "Rambow, Owen and Wanner, Leo and Apidianaki, Marianna and Al-Khalifa, Hend and Eugenio, Barbara Di and Schockaert, Steven",booktitle = "Proceedings of the 31st International Conference on Computational Linguistics",month = jan,year = "2025",address = "Abu Dhabi, UAE",publisher = acl,url = anth # {2025.coling-main.188/},pages = "2765--2780"}
